\documentclass[10pt,twocolumn,letterpaper]{article}

\usepackage[pagenumbers]{cvpr} 

\usepackage[utf8]{inputenc} 
\usepackage[T1]{fontenc}    
\usepackage{url}            
\usepackage{nicefrac}       
\usepackage{microtype}      

\usepackage{wrapfig}
\usepackage{graphicx}
\usepackage{amsmath}
\usepackage{amssymb}
\usepackage{pifont}
\usepackage{booktabs}
\usepackage{array}
\usepackage{graphicx, amsmath, multirow, overpic, textpos} 
\usepackage[dvipsnames]{xcolor}
\usepackage{colortbl}
\usepackage[linesnumbered,ruled,vlined]{algorithm2e}
\SetKwInOut{Parameter}{Parameters}
\usepackage{listings}
\usepackage{tikz}
\usepackage{bm}
\usepackage{arydshln}
\usepackage{hhline}
\usepackage{enumitem}
\usepackage{tabu}
\usepackage{soul}
\usepackage{setspace}
\usepackage{minitoc}

\definecolor{highlight}{HTML}{FDF9D5}
\newlength\savewidth

\newcommand\shlinesmall{\noalign{\global\savewidth\arrayrulewidth
  \global\arrayrulewidth 0.75pt}\hline\noalign{\global\arrayrulewidth\savewidth}}
\newcommand{\tablestyle}[2]{\setlength{\tabcolsep}{#1}\renewcommand{\arraystretch}{#2}\centering\scriptsize}

\renewcommand{\paragraph}[1]{\vspace{+.3mm}\noindent\textbf{#1}}
\newcolumntype{x}[1]{>{\centering\arraybackslash}p{#1pt}}
\newcolumntype{y}[1]{>{\raggedright\arraybackslash}p{#1pt}}
\newcolumntype{z}[1]{>{\raggedleft\arraybackslash}p{#1pt}}
\newcommand{\app}{\raise.17ex\hbox{$\scriptstyle\sim$}}

\definecolor{deemph}{gray}{0.58}
\newcommand{\gc}[1]{\textcolor{deemph}{#1}}
\definecolor{baselinecolor}{gray}{.9}

\newcommand{\gr}{\rowcolor[gray]{.95}}
\newcommand{\grgr}{\rowcolor[gray]{0.9}}
\newcommand{\figref}[1]{Figure~\ref{#1}}
\newcommand{\tabref}[1]{Table~\ref{#1}}

\newcommand{\secref}[1]{Section~\ref{#1}}
\newcommand{\RNum}[1]{\uppercase\expandafter{\romannumeral #1\relax}}

\newcommand*\circled[1]{\tikz[baseline=(char.base)]{
            \node[shape=circle,draw,inner sep=0.5pt] (char) {\textbf{#1}};}}

\definecolor{userborder}{rgb}{0.2608, 0.7071, 0.0218}
\definecolor{brightgreen}{rgb}{0.001, 0.921, 0.38}
\definecolor{darkgreen}{rgb}{0.3608, 0.6471, 0.2118}
\definecolor{userfont}{RGB}{0, 0, 0}
\definecolor{darkred}{rgb}{0.8,0.02,0.02}
\definecolor{ddarkred}{rgb}{0.5,0.02,0.02}
\definecolor{highlightred}{HTML}{ffcbc0}
\definecolor{cvprblue}{rgb}{0.21,0.49,0.74}

\definecolor{cadmiumgreen}{rgb}{0.0, 0.42, 0.24}
\definecolor{aliceblue}{rgb}{0.91, 0.94, 0.97}
\definecolor{darkblue}{rgb}{0.83, 0.89, 0.97}
\definecolor{Blue9}{rgb}{0.098,0.3,0.9}
\definecolor{citecolor}{HTML}{0071bc}

\definecolor{paviolet}{HTML}{DEDBE9}
\definecolor{pasky}{HTML}{D5EBEE}
\definecolor{payellow}{HTML}{FFF6EA}
\definecolor{pared}{HTML}{F9E5ED}
\usepackage[pagebackref,breaklinks,colorlinks]{hyperref}
\hypersetup{
  linkcolor = citecolor,
  citecolor  = citecolor,
  colorlinks = true,
  urlcolor = Blue9
}

\usepackage[most]{tcolorbox}

\def\paperID{38686} 
\def\confName{CVPR}
\def\confYear{2026}
\newcommand{\MSeR}{MM-SeR}

\title{\MSeR: Multimodal Self-Refinement for Lightweight Image Captioning}

\newcommand*{\affmark}[1][*]{\textsuperscript{#1}}

\author{
{Junha Song}\affmark[1],\,\;
{Yongsik Jo}\affmark[2],\,\;
{So Yeon Min}\affmark[3],\,\;
{Quanting Xie}\affmark[3],\,\;
\\
{Taehwan Kim}\affmark[2],\,\;
{Yonatan Bisk}\affmark[3],\,\;
{Jaegul Choo}\affmark[1]\,\;
\\[+.5em]
\affmark[1]KAIST,\;\;
\affmark[2]UNIST,\;\;
\affmark[3]Carnegie Mellon University\\
}

\begin{document}
\twocolumn[{
    \renewcommand\twocolumn[1][]{#1}
    \maketitle
    \begin{center}
        \captionsetup{type=figure}
        \vspace{-1em}
        \includegraphics[width=\textwidth]{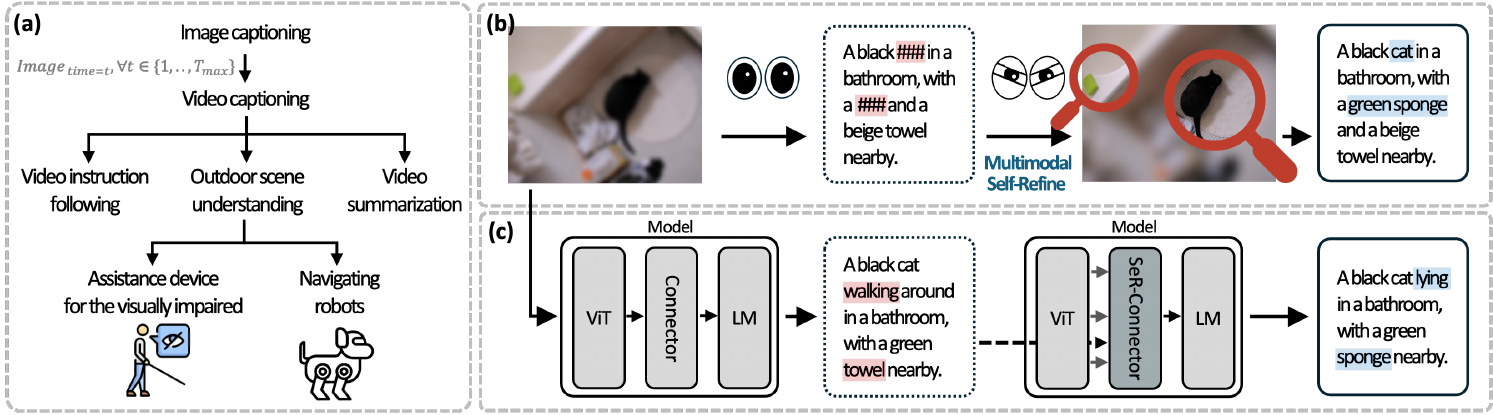}
        \vspace{-1.5em}
        
        \caption{
          (a) Image captioning is a critical component for numerous assistive applications. However, current models often struggle to balance computational efficiency with performance, facing either deployment constraints or limited capability. We introduce a framework inspired by (b) the human visual process, which typically involves perceiving the global scene context before attending to local regions for specific details. (c) This observation highlights the necessity for multimodal self-refinement, a process our framework is designed to perform.
          \vspace{2mm}
        }
        \label{fig:figure1}
    \end{center}
}]

\begin{abstract}

\vspace{-.5em}
Systems such as video chatbots and navigation robots often depend on streaming image captioning to interpret visual inputs. 
Existing approaches typically employ large multimodal language models (MLLMs) for this purpose, but their substantial computational cost hinders practical application.
This limitation motivates our development of a lightweight captioning model. 
Our investigation begins by replacing the large-scale language component in MLLMs with a compact 125M-parameter model.
Surprisingly, this compact model, despite a 93x reduction in size, achieves comparable performance to MLLMs, suggesting that factual image captioning does not significantly require the complex reasoning abilities of LLMs. 
Despite this promising result, our lightweight model still lacks reliability. 
To address this, we draw inspiration from the human visual process: perceiving a global and coarse understanding of the scene before attending to finer details. 
Accordingly, we propose a multimodal self-refinement framework that guides the model to utilize features from salient regions, identified by referencing the previous coarse caption, and to produce a refined description. 
Experimental results demonstrate the superiority of our model in both single-sentence and detailed captioning, extending even to long-range video QA tasks. 
\enlargethispage{1\baselineskip}

\end{abstract}
\section{Introduction}
\label{sec:intro}

Recent progress in image captioning has been driven by the remarkable capability of Multimodal Large Language models (MLLMs)~\cite{llava, li2023blip2}. Building on these advances, image captioning has become a crucial component in various applications. For example, video-based chatbot systems utilize frame-wise \textit{caption} generation for temporal understanding~\cite{wang2023internvid, moment-gpt, videotree}, while navigation robots construct graph-structured scene \textit{descriptions} to operate in complex environments~\cite{xie2024embodied, jiang2024roboexp}. Despite this progress, the substantial computational demands of MLLMs~\cite{jin2024efficient} pose a significant barrier to their practical deployment.

In many industrial systems, detection and segmentation models with fewer than 500M parameters are commonly deployed. This gap led us to ask whether captioning is truly so difficult that it must rely on MLLMs. To address this question, we design a lightweight captioning model that follows the architectural pattern of recent MLLMs and evaluate it on several captioning tasks. Specifically, we implement our model within the LLaVA framework~\cite{llava} by replacing LLaMA-7B~\cite{touvron2023llama} with OPT-125M~\cite{zhang2022opt}, a language model that is 56× smaller in parameter count.
Surprisingly, the resulting model not only competes with MLLMs on the standard MS COCO~\cite{mscoco} benchmark but also performs strongly on more challenging detailed captioning tasks~\cite{dci}, while outperforming existing small captioning models~\cite{ramos2023smallcap, huang2023tag2text}. 
These results reveal the insight that \textbf{the complex capabilities of LLMs are less critical for tasks that focus on enumerating \textit{factual} visual details}. 
These results also suggest that applying modern multimodal architectural designs to small captioning models is required to unlock their potential for deployment in real applications.

Despite the promising performance, the resulting model still exhibits a reliability gap compared to MLLMs. We attribute this gap primarily to visual blindness; prior work \cite{cambrian,eyeswideshut} has highlighted that MLLMs often suffer from ambiguous visual features, which limits their ability to capture fine-grained details. This finding necessitates a method to supply the model with clearer and more informative visual inputs. We address this limitation by adopting a process similar to human visual perception, which begins with a global understanding of the scene before attending to local details. This multi-stage human approach contrasts with conventional captioning models, which typically operate in a single pass, processing the image only once to produce a description.
To this end, \textbf{we design a new framework, Multimodal Self-Refinement (\MSeR)}, which enables the lightweight model to emulate this multi-stage human process. Specifically, \MSeR\ enables the model to first generate an initial caption. The model then leverages this caption to guide its attention toward salient visual regions and extract richer information from the multi-layer features of the vision encoder. This refinement process allows the model to produce a more accurate final description.



In our experiments, we evaluate the \MSeR\ framework on diverse benchmarks, extending beyond standard MS COCO~\cite{mscoco} to include detailed captioning and long-form VideoQA tasks. We compare our model with \MSeR\ against both existing small captioners and MLLMs, demonstrating comparable or even superior performance. 
Particularly in the challenging long-form VideoQA setting, our lightweight model not only outperforms other small specialists but also approaches the accuracy of MLLM generalists. This is accomplished while utilizing \textbf{93\%} fewer parameters and achieving \textbf{82\%} shorter inference time compared to the MLLMs.
These results indicate that the proposed baseline and refinement framework offer a practical route toward lightweight captioning suitable for resource-constrained and on-device applications.
\vspace{+.3em}
\section{Motivation \& Scope}
\label{sec:motivation}
\noindent\textbf{Image Captioning as Foundational Technology.}
Image captioning converts visual content into natural language descriptions~\cite{mokady2021clipcap,hu2022scaling}.
Beyond being a standalone task, it serves as a core component in various vision–language applications.
In video-grounded chatbot systems~\cite{sarto2023positive,wang2024internvideo2,zhao2023learning,moment-gpt}, captioning generates textual representations of multiple frames, which are then integrated into LLMs as prompts to guide instruction following.
Similarly, exploration robots~\cite{xie2024embodied,jiang2024roboexp} operating in disaster environments rely on captioning to encode observed scenes, enabling navigation and human interaction.
In this work, we study image captioning as a key enabler for these applications.

\vspace{+.3em}\noindent\textbf{Real-World Deployment Challenges.}
Recent studies on the above applications often employ open-source MLLMs~\cite{wang2023internvid} or cloud-based APIs (\eg, OpenAI API)~\cite{xie2024embodied} as image captioners.
In practice, these approaches face two major limitations: (1) open-source MLLMs demand computational resources beyond the capacity of edge devices~\cite{cai2020tinytl,song2023ecotta} as shown in \tabref{tab:memory}, and (2) cloud-based APIs rely on stable network connectivity, which may be unavailable in disaster environments.
Moreover, repeated captioning across multiple scenes~\cite{zhang2023simple} further increases the computational burden, making real-world deployment more difficult.

\begin{table}[h]
\tablestyle{.4pt}{1.3} 
\vspace{-.5em}
\caption{Available memory on edge devices and \textbf{GPU usage of recent MLLMs} by parameter size under FP16 precision.}
\vspace{-.5em}
\resizebox{\linewidth}{!}{
\begin{tabular}{x{36} x{36} ;{1.5pt/1.pt} x{60} x{60} x{60}}
\shlinesmall
\multicolumn{2}{c ;{1.5pt/1.pt}}{Edge devices} & LLaVA-1.5 7B,  & LLaVA-NeXT 34B, & LLaVA-OA 72B,  \\[-.3em]
iPhone 16      & Galaxy S25      & mPLUG-Owl3 8B & InternVL 40B   & Qwen2-VL 72B \\
\hdashline
8GB            & 12GB            & 16GB $+$          & 68GB $+$           & 140GB $+$       \\
\shlinesmall
\end{tabular}%
}
\vspace{-.5em}
\label{tab:memory}
\end{table} 
\section{Exploring Lightweight Captioning}
\label{sec:explore}

\begin{table*}[t] 
    \centering
    \caption{\textbf{Comparison of captioning performance.} We evaluate our model and existing captioning models. Despite not introducing any newly proposed methods, our model achieves strong performance. Here, `G' represents a MLLM generalist, while `S' denotes the small captioning model. `*' indicates models that are fine-tuned in this study, as they were not trained for detailed captioning tasks.}
    \vspace{-.5em}
    \resizebox{\linewidth}{!}{%
    \begin{tabular}{x{5} ;{0.2pt/1.5pt} x{83} x{33} ;{0.2pt/1.5pt} x{41} x{41} ;{0.2pt/1.5pt} x{43} x{43} x{43} x{43} ;{0.2pt/1.5pt} x{41}  x{41}}
    \shlinesmall
    \grgr
    \multicolumn{11}{c}{MS COCO~\cite{mscoco}} \\
    & \textbf{model}      & \textbf{venue}                & \textbf{\#\,data}   & \textbf{\#\,params}   & \textbf{B@4~\cite{papineni2002bleu}}    & \textbf{MET~\cite{meteor}}   & \textbf{CIDEr~\cite{vedantam2015cider}}   & \textbf{BERT~\cite{bertscore}}   & \textbf{CLAIR~\cite{clair}}    & \textbf{GPT~\cite{mllmjudge}} \\
    \hdashline
\multirow{5}{*}{\gc{G}} 
    & \gc{InstructBLIP~\cite{dai2023instructblip}}   & \gc{\scriptsize{NeurIPS23}}   & \gc{130M>}             & \gc{8.2B}             & \gc{38.0}       & \gc{29.4}           & \gc{127.8}       & \gc{69.1}              & \gc{-} & \gc{-} \\
    & \gc{Unified-IOXL~\cite{lu2023unifiedio}}   & \gc{\scriptsize{ICLR23}}      & \gc{130M}             & \gc{7.3B}             & \gc{37.0}       & \gc{29.5}           & \gc{123.6}       & \gc{68.2}              & \gc{-} & \gc{-} \\
    & \gc{Shikra~\cite{chen2023shikra}}         & \gc{\scriptsize{arXiv23}}     & \gc{-}             & \gc{7.2B}             & \gc{-}          & \gc{-}           & \gc{117.5}       & \gc{-}              & \gc{-} & \gc{-} \\
    & \gc{Qwen-VL~\cite{bai2023qwenvl}}        & \gc{\scriptsize{arXiv23}}     & \gc{1.5B}             & \gc{9.6B}             & \gc{39.1}          & \gc{30.1}           & \gc{131.9}       & \gc{\textbf{69.8}}              & \gc{77.8{\footnotesize $\pm$3.4}} & \gc{2.89{\footnotesize $\pm$0.11}} \\
    & \gc{LLaVA-1.5~\cite{llava15}}      & \gc{\scriptsize{CVPR24}}      & \gc{1M}             & \gc{7.3B}             & \gc{39.4}       & \gc{29.5}           & \gc{133.7}       & \gc{69.4}              & \gc{78.1{\footnotesize $\pm$3.8}} & \gc{2.93{\footnotesize $\pm$0.10}}\\
    & \gc{Cambrian~\cite{cambrian}}      & \gc{\scriptsize{NeurIPS24}}      & \gc{2M}             & \gc{10.5B}      & \gc{\textbf{40.1}}  & \gc{\textbf{30.9}}   & \gc{\textbf{137.5}}   & -    & \gc{\textbf{78.2}{\footnotesize $\pm$3.2}} & \gc{\textbf{3.02}{\footnotesize $\pm$0.13}}\\
\hdashline
    & I-Tuning~\cite{luo2023ituningtuningfrozenlanguage}             & \scriptsize{ICASSP23}         & 0.5M                  & 250M                  & 35.5            & 28.8                & 120.0            & -                   & -   & 2.50{\footnotesize $\pm$0.10}\\
    & CapPa~\cite{tschannen2023image}               & \scriptsize{NeurIPS23}        & 1B                  & 650M                  & -               & -                & 125.8            & -                   & -  & 2.67{\footnotesize $\pm$0.08}\\
    & LocCa~\cite{wan2024locca}               & \scriptsize{NeurIPS24}        & 1B                 & 600M                  & -               & -               & 127.1            & -                  & - & 2.66{\footnotesize $\pm$0.11}  \\
    & SmallCap~\cite{ramos2023smallcap}             & \scriptsize{CVPR23}           & 0.5M                  & 450M                  & 37.6            & 28.7                & 122.7            & 67.2                   & 73.7{\footnotesize $\pm$3.9}  & 2.46{\footnotesize $\pm$0.10}\\
    & Tag2Text~\cite{huang2023tag2text}            & \scriptsize{ICLR24}           & 4M                 & 900M                  & 38.4            & 30.0               & 128.7            & 69.3                  & 76.1{\footnotesize $\pm$3.1}  & \textbf{2.78}{\footnotesize $\pm$0.08} \\
    & ViPCap~\cite{vipcap}            & \scriptsize{AAAI25}           & 4M                 & 225M                  & 37.7            & 28.6               & 122.9            & -                 & -  & - \\
\gr
\multirow{-6}{*}{\cellcolor{white}S}   & Ours      &                               & 1M                 & 450M                  & \textbf{39.4}            & \textbf{30.3}               & \textbf{129.6}            & \textbf{69.4}                  & \textbf{76.3}{\footnotesize $\pm$3.0}  & 2.74{\footnotesize $\pm$0.06}\\
    \shlinesmall
    \multicolumn{11}{c}{}\\[-.4em]
    \shlinesmall
    \grgr
    \multicolumn{11}{c}{ShareGPT4V~\cite{chen2024sharegpt4v} \& DCI~\cite{dci}} \\
                                       & \textbf{model}      & \textbf{venue}                & \textbf{\#\,data}   & \textbf{\#\,params}   & \textbf{B@4~\cite{papineni2002bleu}}    & \textbf{CIDEr~\cite{vedantam2015cider}}    & \textbf{BERT~\cite{bertscore}} & \textbf{CAPT~\cite{dong2024benchmarking}}   & \textbf{CLAIR~\cite{clair}}    & \textbf{GPT~\cite{mllmjudge}} \\
    \hdashline
\multirow{3}{*}{\gc{G}} 
    & \gc{Qwen-VL~\cite{bai2023qwenvl}} & \gc{\scriptsize{arXiv23}}  & \gc{1.5B}          & \gc{9.6B}   & \gc{10.8}  & \gc{35.6}            & \gc{37.2}  & \gc{48.4}                & \gc{57.5{\footnotesize $\pm$3.2}}  & \gc{\textbf{3.05}{\footnotesize $\pm$0.08}}\\
    & \gc{LLaVA-1.5~\cite{llava15}}      & \gc{\scriptsize{CVPR24}}         & \gc{1M}     & \gc{7.3B}   & \gc{10.6}           & \gc{36.1}   & \gc{36.6}           & \gc{48.6}       & \gc{  {\footnotesize $\pm$  }}           & \gc{  {\footnotesize $\pm$  }}\\
    & \gc{EyesWideShut~\cite{eyeswideshut}} & \gc{\scriptsize{CVPR24}}      & \gc{1M}     & \gc{7.6B}   & \gc{11.6}           & \gc{36.5}   & \gc{37.2}           & \gc{49.0}       & -                                           & -                                 \\
    & \gc{Cambrian~\cite{cambrian}}         & \gc{\scriptsize{NeurIPS24}}   & \gc{2M}     & \gc{10.5B}  & \gc{\textbf{13.3}}           & \gc{\textbf{38.7}}   & \gc{\textbf{38.1}}           & \gc{50.1}       & \gc{\textbf{58.2}{\footnotesize $\pm$3.1}}           & \gc{3.03{\footnotesize $\pm$0.09}}\\
\hdashline
    & SmallCap*~\cite{ramos2023smallcap}           & \scriptsize{CVPR23}           & 0.5M                 & 450M                  & 14.5             & 20.1              & 28.9                & 23.3            & -     & - \\
    & Tag2Text*~\cite{huang2023tag2text}           & \scriptsize{ICLR24}           & 4M                 & 900M                  & 17.8             & 32.5              & 40.7                & 40.1            & 54.2{\footnotesize $\pm$3.1}  & 2.72{\footnotesize $\pm$0.14} \\
\gr
\multirow{-3}{*}{\cellcolor{white}S}   & Ours      &                               & 1M                 & 450M                  & \textbf{18.0}             & \textbf{40.5}              & \textbf{43.1}                & \textbf{45.9}            & \textbf{54.6}{\footnotesize $\pm$3.4}   & \textbf{2.74}{\footnotesize $\pm$0.12}  \\
    \shlinesmall
    \multicolumn{11}{c}{}\\[-.4em]
    \shlinesmall
    \grgr
    \multicolumn{11}{c}{GLaMM~\cite{rasheed2024glamm}} \\
                                       & \textbf{model}      & \textbf{venue}                & \textbf{\#\,data}   & \textbf{\#\,params}   & \textbf{B@4~\cite{papineni2002bleu}}    & \textbf{CIDEr~\cite{vedantam2015cider}}    & \textbf{BERT~\cite{bertscore}} & \textbf{CAPT~\cite{dong2024benchmarking}}   & \textbf{CLAIR~\cite{clair}}    & \textbf{GPT~\cite{mllmjudge}} \\
    \hdashline
\multirow{1}{*}{\gc{G}} 
    & \gc{LLaVA-1.5~\cite{llava15}}      & \gc{\scriptsize{CVPR24}}      & \gc{1M}            & \gc{7.3B}             & \gc{8.8}        & \gc{23.4}          & \gc{35.1}           & \gc{40.0}       & \gc{53.8{\footnotesize $\pm$4.0}}    & \gc{3.02{\footnotesize $\pm$0.10}}  \\
\hdashline
\gr
\multirow{1}{*}{\cellcolor{white}S}    & Ours      &                               & 1M                 & 450M                  & \textbf{16.5}             & \textbf{29.1}              & \textbf{38.7}                & \textbf{42.0}            & \textbf{51.8}{\footnotesize $\pm$4.1}     & \textbf{2.64}{\footnotesize $\pm$0.09} \\
    \shlinesmall
    \end{tabular}%
    }
    \vspace{-1.em}
    \label{tab:gen_perform}
\end{table*}

Motivated by the challenges discussed above, we explore a lightweight captioning model and examine its performance through extensive evaluation.

\vspace{+.3em}\noindent\textbf{Model construction.}
To construct a lightweight captioning model, we aim to reduce the dependence on LLMs, which account for most of the computational cost in MLLMs (\eg, 96\% in LLaVA-7B~\cite{llava} arises from LLaMA~\cite{touvron2023llama}).
Accordingly, we replace the LLaMA-7B in LLaVA-1.5 with OPT-125M~\cite{zhang2022opt}, a 56× smaller language model.

\vspace{+.3em}\noindent\textbf{Experimental details.}
We adopt the publicly available LLaVA-1.5~\cite{llava15} codebase. Except for replacing the language model, all training configurations remain consistent with the original setup, including batch size and learning rate.
More implementation details are provided in \secref{sec:added_implementation}.
and our source code\footnotemark.
To train our model, we first pretrain the multimodal connector on the Caption Concept-balanced 558K dataset~\cite{llava15}, followed by fine-tuning on task-specific datasets such as MS COCO~\cite{mscoco}, DCI~\cite{dci}, and ShareGPT4V~\cite{chen2024sharegpt4v}.
We evaluate our model using standard metrics, BLEU~\cite{papineni2002bleu}, CIDEr~\cite{vedantam2015cider}, and BERTScore~\cite{bertscore}, as well as MLLM-as-a-Judge~\cite{mllmjudge,clair} with GPT-4o-mini~\cite{achiam2023gpt}.
\footnotetext{\url{https://github.com/junha1125/Lightweight-Captioner}\vspace{-2em}}

\vspace{+.3em}\noindent\textbf{Generalist vs. Specialist.} 
Following prior studies~\cite{ye2024mplug,mome,wu2024visionllm}, we define generalists as MLLMs trained on diverse datasets for broad objectives and evaluated in a zero-shot manner using publicly released weights.
In contrast, specialists refer to compact captioning models such as SmallCap~\cite{ramos2023smallcap} and Tag2Text~\cite{huang2023tag2text}, which are trained and optimized exclusively on task-specific data.

\subsection{Results on Single Sentence Captioning}
\begin{figure}[t]\centering
\vspace{+.5em}
\includegraphics[width=1.0\linewidth]{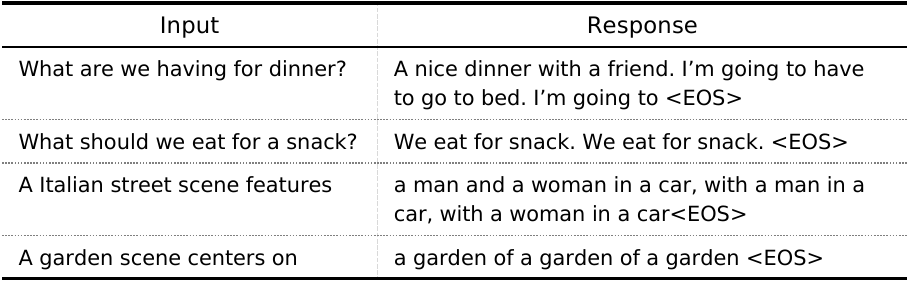}
\vspace{-1.5em}
\caption{\textbf{Examples generated by OPT-125M} show its limited capacity. However, we found that with slight fine-tuning, the model exhibits a surprisingly strong captioning ability.}
\vspace{-1.5em}
\label{fig:otp125}
\end{figure}

We use the MS COCO Captions dataset~\cite{mscoco}, the most widely adopted benchmark for image captioning, where each caption consists of a single sentence, with an average length of about ten words.
As shown in \tabref{tab:gen_perform}, our model outperforms previous small-scale captioning methods~\cite{huang2023tag2text,wan2024locca} that contain fewer than 1B parameters.
In particular, it achieves a CIDEr score 6.9 points higher than SmallCap~\cite{ramos2023smallcap}, which also employs OPT-125M as its language backbone.
Despite the absence of any newly introduced techniques, the model demonstrates unexpectedly strong performance, which we further analyze in \secref{sec:whyexisting}.
Moreover, when compared with generalist MLLMs~\cite{chen2023pali,bai2023qwenvl}, our model attains comparable results while requiring far less computational resources.


\begin{figure*}[t]\centering
\includegraphics[width=\linewidth]{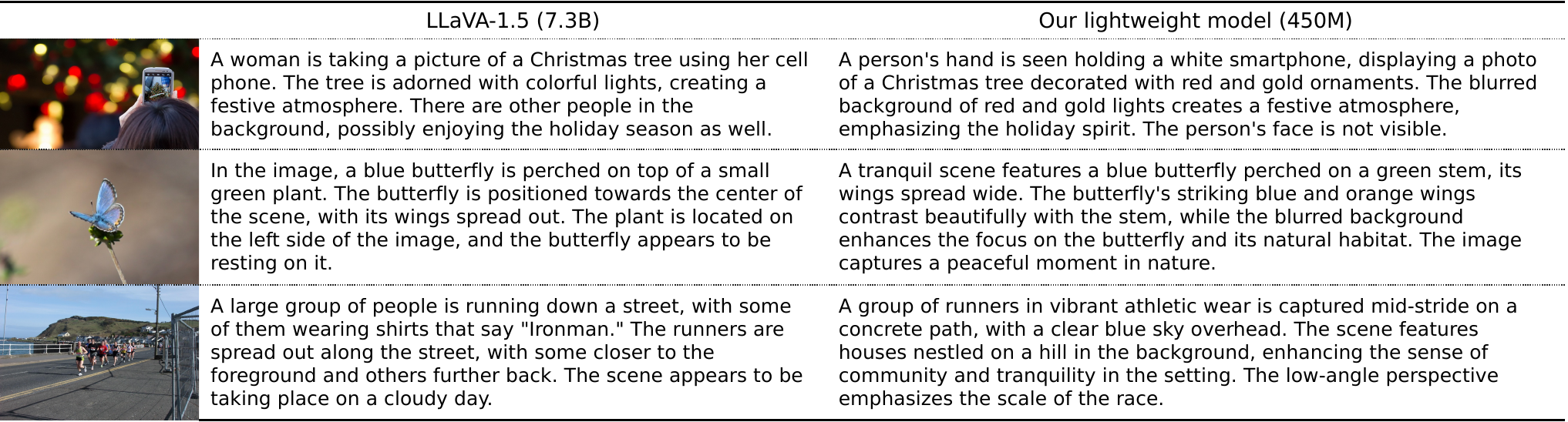}
\vspace{-1.7em}
\caption{\textbf{Qualitative results on detailed captioning.} 
Despite its small size, our lightweight specialist produces captions that are comparable in quality to those generated by large-scale MLLMs.  
These results highlight that effective detailed captioning are achievable even with a lightweight language model.  
Qualitative results across various MLLM models are also presented in \figref{fig:gen_quali2}, \figref{fig:limitation_mllm}, and \figref{fig:limitation}.}
\label{fig:gen_quali}
\vspace{-1.3em}
\end{figure*}



\subsection{Results on Detailed Captioning}

For this experiment, we fine-tune our model on the DCI~\cite{dci} and ShareGPT4V~\cite{chen2024sharegpt4v} datasets, and additionally include GLaMM~\cite{rasheed2024glamm}.
We initially expected the lightweight model to underperform on this task, as OPT-125M offers limited language capacity due to its small parameter size, as shown in \tabref{fig:otp125}.
However, the results in \tabref{tab:gen_perform} and \figref{fig:gen_quali} show that our model achieves unexpectedly strong performance on detailed captioning, contrary to our assumption.



\subsection{Key Insight} 

Although MLLMs depend on the advanced reasoning capabilities of large language models for tasks such as visual question answering and instruction following, our experiments reveal a contrasting trend in factual image captioning.
We find that accurate caption generation can be achieved without such reasoning-intensive processes, as a lightweight 125M-parameter language model attains comparable performance to full-scale MLLMs.
This result implies that image captioning primarily relies on perceptual grounding rather than abstract reasoning, suggesting that \ul{a compact model can serve as an efficient and practical alternative for captioning-based applications}.

\vspace{+1.em}
\section{\MSeR: Multimodal Self-Refinement} %
\label{sec:glance}

\begin{figure*}[t]\centering
\includegraphics[width=\linewidth]{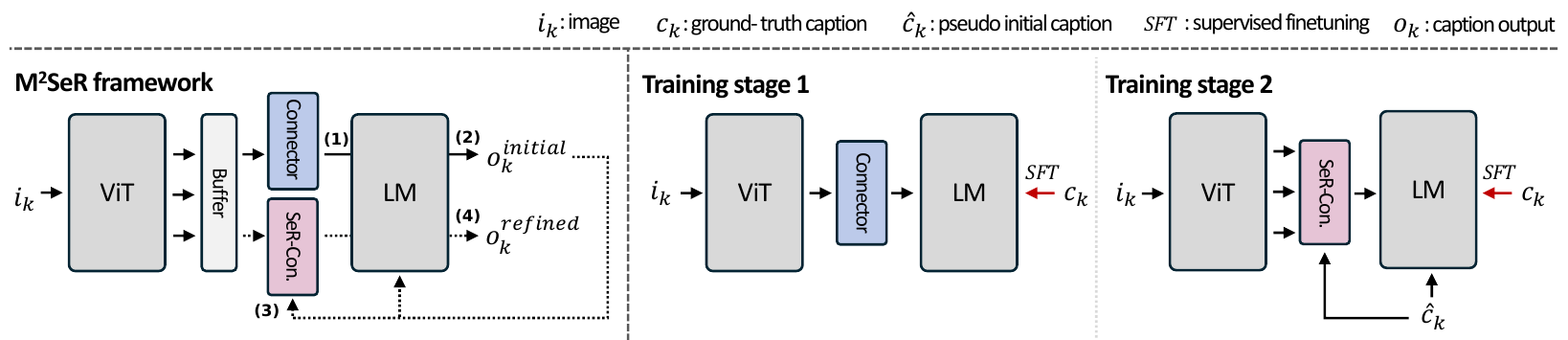}
\vspace{-2em}
\caption{
\textbf{Overview of the Multimodal Self-Refinement (\MSeR) framework.}
(left) The model first generates an initial caption, which guides the extraction of refined visual features for a second-stage generation.
Fine-tuning is performed in two stages (right): (1) supervised training using ground-truth captions, and (2) refinement training using pseudo-initial captions that slightly deviate from the ground-truth to encourage self-correction. Here, `SeR-Con.' denotes the SeR-Connector, which differs from the connector used for generating the initial caption. It processes additional inputs and is utilized in the refinement process.}
\label{fig:workflow}
\vspace{-1.3em}
\end{figure*}

Despite the impressive performance of our lightweight captioner, we still observe a capacity gap relative to large-scale MLLMs in certain evaluations.
To address this limitation, we introduce a new framework, \textit{Multimodal Self-Refinement} (\MSeR), illustrated in \figref{fig:figure1}, which improves caption quality through self-guided refinement.
First, we draw from the human description process: forming an initial global understanding before attending to salient details.
Following this principle, \MSeR\ adopts a multi-stage generation procedure.
The model first produces an initial caption capturing the overall scene, and then uses this possibly coarse output to guide the extraction of clearer and more informative visual features, which support subsequent refinement. 

\subsection{Proposed Framework}
\label{sec:framework}

As illustrated in \figref{fig:workflow}, our framework extends the conventional single-pass captioning approach by incorporating a self-refinement stage.
Although the same language model is used for both the initial caption and the refinement step, the refinement process operates with a dedicated connector (\ie, SeR-Connector). This connector is specifically tasked with integrating the following inputs: the previously generated caption and multi-layer features from the vision encoder.
These inputs each serve a distinct purpose, as detailed below.

\vspace{+.5em}\noindent\textbf{Looking at what matters.}
When asked to refine a caption such as “A cat relaxing on a brown chair,” humans naturally attend to the key elements referenced in the text, like the cat and the chair. 
Following this intuition, we feed the initial caption into the SeR-Connector and Language models, allowing the modules to identify visually relevant regions and direct their attention toward them during refinement.

\vspace{+.5em}\noindent\textbf{Looking in detail.}
Coarse visual features often limit fine-grained description quality.
While some prior studies~\cite{cambrian,eyeswideshut,fromcliptodino} address this by adding auxiliary vision encoders, this strategy increases model size; for instance, Interleaved MoF~\cite{eyeswideshut} introduces DINOv2~\cite{oquab2024dinov}, adding 300M parameters—a 66.7\% increase relative to our 450M-parameter model.
Instead of expanding the architecture, we aim to maximize the utility of the existing encoder by leveraging multi-layer features, which provide richer and more detailed representations.
Although earlier works~\cite{fpn} have explored multi-layer features, our approach uses them specifically to supply finer-grained visual cues for the refinement stage.



\vspace{+.5em}\noindent\textbf{Novelty of \MSeR.}
This refinement paradigm parallels the concept of \textit{self-refinement} explored in LLM research~\cite{selfrefine,le2022coderl,pan2023automatically}.
By extending this idea to the multimodal domain, our framework integrates visual evidence directly into the refinement, to the best of our knowledge, the first attempt to realize self-refinement within multimodal models.

\subsection{Training Strategy}
\label{sec:training}

For the proposed framework to operate effectively, the model needs to be trained to produce reliable initial captions and to extract the refined visual features described in \figref{fig:workflow}. To achieve this, we adopt a two-stage training strategy.

In the first stage, the model is trained to generate the initial caption following the standard LLaVA procedure~\cite{llava15}. 
In the second stage, the model learns how to perform refinement.
Let the training set be \( X = \{x_1, x_2, \dots, x_N\} \), where each \( x_k = (i_k, c_k) \) contains an image and its ground-truth caption.
A straightforward strategy might treat the model’s momentary first-pass caption as the refinement input and the corresponding ground-truth caption \(c_k\) as the target.
However, this setup often produces pairs with little semantic alignment.
For example, if the initial caption is ``a table in front of a window’’ while the \(c_k\) is ``a cat sitting on a table,’’ the two descriptions offer no meaningful basis for progressive refinement.
Training on such mismatched pairs would likely lead the model to disregard the initial caption and simply regenerate a new one, rather than learn how to refine it. Further discussion is in \secref{sec:peudo_caption}.

To address this issue, we generate pseudo-initial captions \(\hat{c}_{k}\) by prompting GPT-4o-mini~\cite{achiam2023gpt} to introduce small perturbations to entities, attributes, or relations in the \(c_k\).  
For example, given the \(c_k\) ``a cat sitting on a chair,’’ the pseudo-initial \(\hat{c}_{k}\) version may become ``a dog sitting on a chair.’’  
During training, the SeR-Connector receives multi-layer visual features together with the \(\hat{c}_{k}\) and learns to extract features that are more informative for caption refinement.
The language model takes the visual embeddings and \(\hat{c}_{k}\) as an additional textual prompt and predicts a refined caption, which is supervised to match \(c_k\).

\vspace{+1.em}\noindent\textbf{Rationale for refinement training.}
Let $\pi_\theta$ denote the language model. Each pseudo‑initial caption $\hat{c}_k$ deviates from the ground truth $c_k$ at only a few token positions $E_k=\{t\mid \hat{c}_{k,t}\neq c_{k,t}\}$. Under the sequence‑level objective

\vspace{-1.2em}
\begin{equation}
    \mathcal{L}(\theta)\;=\;-\!\mathbb{E}\Bigl[\sum_{j}\log\pi_\theta\!\bigl(c_{k,j}\mid i_k,\hat{c}_k,c_{k,<j}\bigr)\Bigr],    
\end{equation}
\vspace{-1.4em}

\noindent
gradients are likely to be primarily concentrated on tokens in $E_k$, leading to a form of \emph{targeted optimization}, in which the model retains the correct parts of $\hat{c}_k$ while rewriting only the erroneous ones. Writing
$\Delta_k(\theta)=\log\pi_\theta(c_k\mid i_k,\hat{c}_k)-\log\pi_\theta(\hat{c}_k\mid i_k,\hat{c}_k),$
we have $\mathcal{L}(\theta)\propto-\mathbb{E}[\Delta_k]$; hence, minimizing $\mathcal{L}$ is equivalent to maximizing the expected margin $\Delta_k$, thereby directly increasing the likelihood of the refined caption relative to its \textit{flawed} precursor.
In effect, each gradient step encourages SeR-Connector to focus on visual regions likely responsible for errors in the initial caption and guides the language model to better interpret these refined features, resulting in more accurate captions.
This targeted optimization resembles the philosophy of Direct Preference Optimization (DPO)~\cite{dpo}, if we consider $c_k$ and $\hat{c}_k$ as preferred and less-preferred responses. In contrast to DPO, which treats both responses symmetrically during optimization, our method offers a new perspective by assigning them distinct roles as input and target.

\begin{figure}[t]\centering
\includegraphics[width=1.\linewidth]{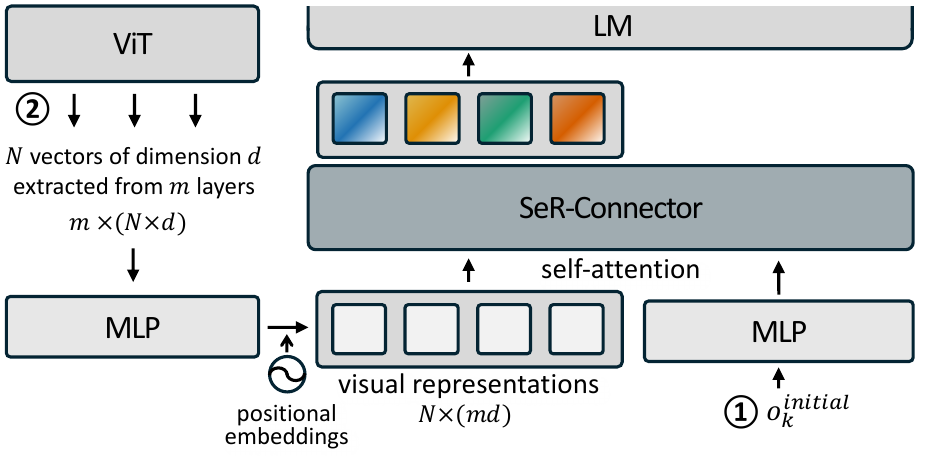}
\vspace{-2em}
\caption{\textbf{Details of the SeR-Connector.}
Unlike the standard two-layer MLP connectors used in typical MLLMs, the SeR-Connector is designed to support the refinement process and to effectively incorporate the inputs described in \secref{sec:framework}.}
\label{fig:ablation_models}
\vspace{-1.5em}
\end{figure}

\section{Experiments}
\label{sec:result}

\begin{table*}[t]
    \centering
    \caption{\textbf{Quantitative results of \MSeR.}
    The results demonstrate the effectiveness of extending single-pass captioning with a self-refinement stage.
    Refinement relies on two key inputs: \ding{172} the initially generated caption and \ding{173} multi-layer features from the vision encoder.
    \MSeR\ yields consistent performance gains on single-sentence and detailed captioning tasks.}
    \vspace{-.5em}
\resizebox{\linewidth}{!}{%
    \begin{tabular}{y{90} x{55} ;{0.3pt/1.5pt} x{52} x{28} ;{0.3pt/1.5pt} x{52} x{28} ;{0.3pt/1.5pt} x{55} x{28} ;{0.3pt/1.5pt} x{55} x{28}}
    \shlinesmall
    \grgr \multicolumn{10}{c}{MS COCO~\cite{mscoco}} \\
    \textbf{model} & \textbf{\#params} & \textbf{B@4~\cite{papineni2002bleu}} & \textbf{gain} & \textbf{CIDEr~\cite{vedantam2015cider}} & \textbf{gain} & \textbf{CLAIR~\cite{clair}} & \textbf{gain} & \textbf{GPT~\cite{mllmjudge}} & \textbf{gain} \\
    \hdashline
    \gc{LLaVA-1.5~\cite{llava15}}                                   & \gc{7.3B} & \gc{39.4} & \gc{-} & \gc{133.7} & \gc{-} & \gc{78.1{\footnotesize $\pm$3.8}} & \gc{-} & \gc{2.93{\footnotesize $\pm$0.10}} & \gc{-} \\
        \hdashline
    Our model                                                  & 450M & 39.4 & - & 129.6 & - & 76.3{\footnotesize $\pm$3.0} & - & 2.74{\footnotesize $\pm$0.06} & - \\
    \gr
    {\scriptsize +} \MSeR\ with \scriptsize{\circled{1} + \circled{2}} & 500M & 39.9 & +0.5 & 133.5 & +3.9 & 77.6{\footnotesize $\pm$2.9} & +1.3 & 2.82{\footnotesize $\pm$0.09} & +0.08 \\
        \hdashline
    {\scriptsize +} \MSeR\ with \scriptsize{\circled{1}}               & 500M & 39.6 & +0.2 & 131.9 & +2.3 & - & - & - & - \\
    {\scriptsize +} \MSeR\ with \scriptsize{\circled{2}}               & 500M & 39.6 & +0.2 & 132.3 & +2.7 & - & - & - & - \\
    Single pass with \scriptsize{\circled{2}}                     & 450M & 39.7 & +0.3 & 130.6 & +1.0 & - & - & - & - \\
    \shlinesmall

    \\[-.6em]
    \shlinesmall
    \grgr \multicolumn{10}{c}{ShareGPT4V~\cite{chen2024sharegpt4v} \& DCI~\cite{dci}} \\
    \textbf{model} & \textbf{\#params} & \textbf{CIDEr~\cite{vedantam2015cider}} & \textbf{gain} & \textbf{CAPT~\cite{dong2024benchmarking}} & \textbf{gain} & \textbf{CLAIR~\cite{clair}} & \textbf{gain} & \textbf{GPT~\cite{mllmjudge}} & \textbf{gain} \\
    \hdashline
    \gc{Cambrian~\cite{cambrian}}                                    & \gc{10.5B} & \gc{38.7} & \gc{-} & \gc{50.1} & \gc{-} & \gc{58.2{\footnotesize $\pm$3.1}} & \gc{-} & \gc{3.00{\footnotesize $\pm$0.10}} & \gc{-} \\
        \hdashline
    Our model                                                  & 450M & 40.5 & - & 45.9 & - & 54.6{\footnotesize $\pm$3.4} & - & 2.74{\footnotesize $\pm$0.12} & - \\
    \gr
    {\scriptsize +} \MSeR\ with \scriptsize{\circled{1} + \circled{2}} & 500M & 43.6 & +3.1 & 48.4 & +2.5 & 57.7{\footnotesize $\pm$3.0} & +3.1 & 3.02{\footnotesize $\pm$0.12} & +0.28 \\
        \hdashline
    {\scriptsize +} \MSeR\ with \scriptsize{\circled{1}}               & 500M & 42.8 & +2.3 & 47.1 & +1.2 & 55.8{\footnotesize $\pm$3.1} & +1.2 & 2.78{\footnotesize $\pm$0.11}  & +0.04 \\
    {\scriptsize +} \MSeR\ with \scriptsize{\circled{2}}               & 500M & 43.1 & +2.6 & 47.6 & +1.7 & 56.8{\footnotesize $\pm$3.4} & +2.2 & 2.88{\footnotesize $\pm$0.12}  & +0.14 \\
    Single pass with \scriptsize{\circled{2}}                     & 450M & 42.5 & +2.0 & 46.5 & +0.6 & 56.3{\footnotesize $\pm$2.9} & +1.7 & 2.90{\footnotesize $\pm$0.11}  & +0.16 \\
    \shlinesmall

    \\[-.6em]
    \shlinesmall
    \grgr \multicolumn{10}{c}{GLaMM~\cite{rasheed2024glamm}} \\
    \textbf{model} & \textbf{\#params} & \textbf{CIDEr~\cite{vedantam2015cider}} & \textbf{gain} & \textbf{CAPT~\cite{dong2024benchmarking}} & \textbf{gain} & \textbf{CLAIR~\cite{clair}} & \textbf{gain} & \textbf{GPT~\cite{mllmjudge}} & \textbf{gain} \\
    \hdashline
    \gc{LLaVA-1.5~\cite{llava15}}                                   & \gc{7.3B} & \gc{23.4} & \gc{-} & \gc{40.0} & \gc{-} & \gc{53.8{\footnotesize $\pm$4.0}} & \gc{-} & \gc{3.02{\footnotesize $\pm$0.10}} & \gc{-} \\
        \hdashline
    Our model                                                  & 450M & 29.1 & - & 42.0 & - & 51.8{\footnotesize $\pm$4.1} & - & 2.64{\footnotesize $\pm$0.09} & - \\
    \gr
    {\scriptsize +} \MSeR\ with \scriptsize{\circled{1} + \circled{2}} & 500M & 30.4 & +1.3 & 42.8 & +0.8 & 53.4{\footnotesize $\pm$3.8} & +1.6 & 2.88{\footnotesize $\pm$0.11} & +0.24 \\
    \shlinesmall
    \end{tabular}%
    }
    \label{tab:ref_perform}
\end{table*}

\begin{figure*}[t]\centering
\includegraphics[width=\linewidth]{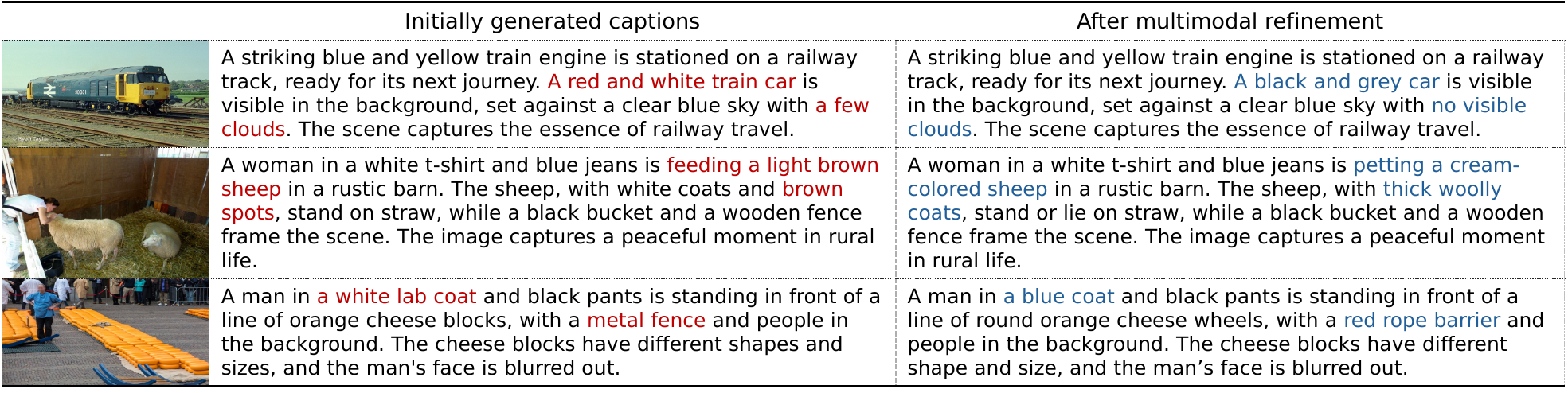}
\vspace{-1.5em}
\caption{\textbf{Qualitative comparison of initial and refined captions.}
The examples demonstrate how the proposed \MSeR\ improves the descriptive quality of image captions.  
Through the refinement process, some entity- and attribute-level errors are corrected, and vague expressions are replaced with more specific and visually grounded descriptions.  
More results can be found in \secref{sec:added_refinement}.}
\label{fig:ref_quali}
\vspace{-1.0em}
\end{figure*}

\subsection{Implementation Details}
We train the framework in two stages.
The first stage runs for 10 epochs to learn initial caption generation, and the second stage runs for 2 epochs to train the refinement process.
Both stages use a batch size of 256 and a learning rate of $2*10^{-5}$.
All experiments are conducted on two NVIDIA A6000 GPUs.
Additional training settings and hyperparameters are described in \secref{sec:added_implementation}.

\vspace{+.5em}\noindent\textbf{SeR-Connector.}
The SeR-Connector is simply implemented with a set of Transformer blocks~\cite{bert}.
As shown in \figref{fig:deeplens}, it receives two inputs: \ding{172} the previously generated caption, encoded as token embeddings, and \ding{173} multi-layer features from the vision encoder.
From the vision encoder, we collect $N$ visual token vectors of dimension $d$ from $m$ selected layers.
The features are concatenated along the channel dimension, yielding a representation of size $N*(md)$ that preserves hierarchical visual information.
The resulting output is then forwarded to the language model for refinement.
The ablation study is discussed in \secref{sec:deeplens_ablation}.

\subsection{Results}
\label{sec:results}

\vspace{+.2em}\noindent\textbf{Effect of \MSeR.} 
The results in \tabref{tab:ref_perform} and \figref{fig:ref_quali} demonstrate the effectiveness of our framework in improving caption quality.
We present ablation studies on the two key inputs of our framework, \ding{172} the initial captions and \ding{173} the multi-layer visual features, to examine their individual contributions.
The refinement stage improves the initial captions by +3.9 and +2.5 CIDEr points for the single-sentence and detailed captioning tasks, respectively.
We further evaluate the use of \ding{173} without applying the refinement stage, observing that single-pass captioning provides only limited gains, indicating the necessity of the step.
In our framework, adding the refinement stage inevitably introduces additional overhead, including roughly 50M parameters for the SeR-Connector and one extra inference step with the language model.
Nevertheless, as shown in \tabref{tab:time}, the computational overhead remains minimal compared with MLLMs.

\begin{figure*}[t]\centering
\includegraphics[width=.9\linewidth]{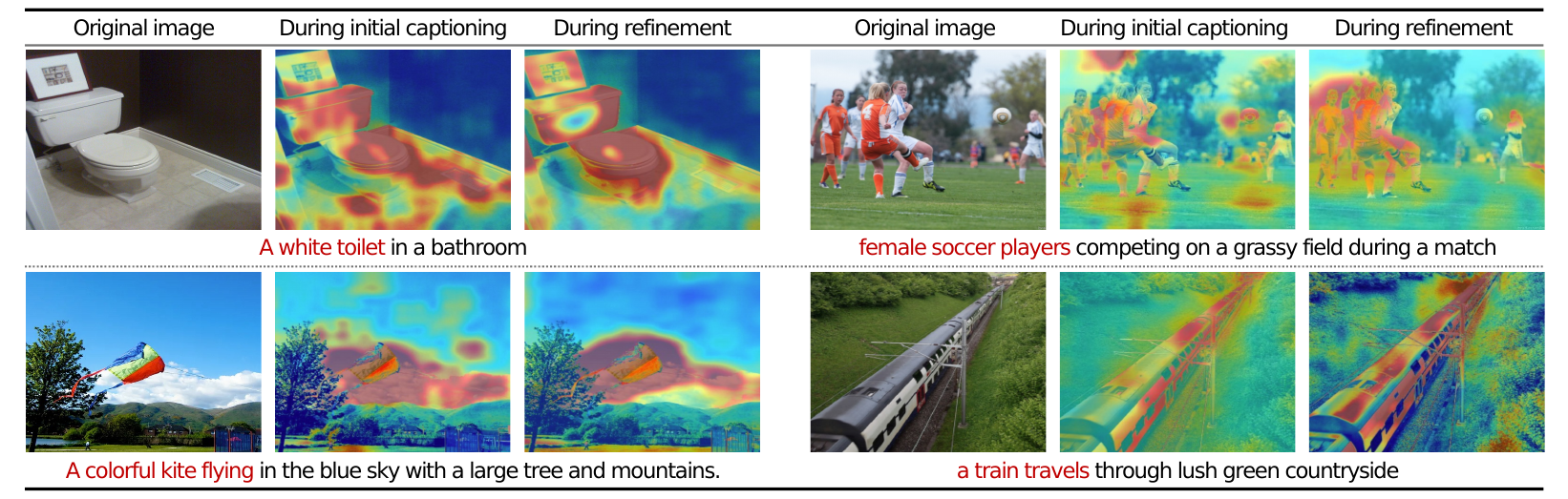}
\vspace{-.5em}
\caption{\textbf{Attention patterns during caption generation.}
The left panel shows the model’s attention when producing captions from a \textit{first-pass view}, where attention is broadly distributed.
The right panel visualizes the refinement stage, in which the initial caption guides the model to focus on more relevant regions associated with the highlighted words.}
\label{fig:attention}
\vspace{-.5em}
\end{figure*}

\begin{table*}[t]
\caption{\textbf{Evaluation on Long-Range VideoQA.}
We follow the baseline setup of LLoVi~\cite{zhang2023simple}, replacing the captioner with our model or MLLMs.
Frame-level captions are aggregated and provided to Qwen2.5-14B for answering video-related questions.
Our specialist delivers competitive accuracy with significantly fewer parameters, demonstrating its applicability even in long-range video understanding.}
\vspace{-.5em}
\tablestyle{1.5pt}{1.1}
\newcolumntype{g}{>{\columncolor[gray]{0.95}} x{50}}
\resizebox{\linewidth}{!}{%
\begin{tabular}{
    x{53} | x{50} ;{.5pt/1.pt} x{45}  x{45} x{45} ;{.5pt/1.pt} x{45} x{45}  g g
}
\shlinesmall
\textbf{LLM} & \textbf{Captioner} & LLaVA-1.0~\cite{llava} & BLIP-2~\cite{li2023blip2} & LLavA-1.5~\cite{llava15} & SmallCap*~\cite{ramos2023smallcap} & Tag2text*~\cite{huang2023tag2text} & Our specialist & {\scriptsize +} \MSeR\ \\
\hdashline
\multirow{3}{*}{Qwen2.5 14B~\cite{yang2024qwen2}} & \textbf{\#params} & 7.3B & 7.4B & 7.3B & 450M & 900M & 450M & 500M \\
& \textbf{accuracy} & 47.6 & 50.6 & \textbf{51.1} & 41.8 & 47.1 & 49.3 & \textbf{50.8}\\
& \textbf{time} & 29m\,20s & 29m\,44s & 29m\,20s & 4m\,45s & 7m\,14s & 4m\,53s & 5m\,10s \\
\shlinesmall
\end{tabular}%
}
\label{tab:longvqa}
\vspace{-1.5em}
\end{table*}




\begin{table}[t]
\tablestyle{1.pt}{1.4}
\caption{Inference time required to generate captions for 100 streaming images. Given the captioning performance in \tabref{tab:ref_perform}, our model demonstrates remarkable efficiency.}
\vspace{-.5em}
\newcolumntype{g}{>{\columncolor[gray]{0.95}} x{70}}
\begin{tabular}{x{30} | x{60} g g}
\shlinesmall
& \textbf{LLaVA-1.5}~\cite{llava15} & \textbf{Ours} & \textbf{+ \MSeR} \\
\hdashline
\textbf{time} 
& 274.49\,s
& 5.55\,s {\scriptsize (97.97\%$\downarrow$)}
& 7.44\,s {\scriptsize (97.28\%$\downarrow$)} \\
\shlinesmall
\end{tabular}
\vspace{-2em}
\label{tab:time}
\end{table}


\vspace{+.2em}\noindent\textbf{Evaluation on long-range video question answering.}
Since captioning models serve as core components in downstream applications, as discussed in \secref{sec:motivation}, we assess their utility in a practical setting by using our captioner as the captioning baseline for the long-range VideoQA task introduced in LLoVi~\cite{zhang2023simple}.
{\small (\textit{Setup:})} For fair comparison, all small specialists~\cite{ramos2023smallcap,huang2023tag2text}, including ours, are trained on ShareGPT4V~\cite{chen2024sharegpt4v} and DCI~\cite{dci}.
Following~\cite{zhang2023simple}, we adopt Qwen2.5-14B as the backbone LLM for answer generation across all captioners to isolate the effect of caption quality.
Inference time is measured end-to-end, accounting for both caption generation and LLM reasoning under a 10-minute video scenario.
{\small (\textit{Results:})} 
As shown in \tabref{tab:longvqa}, our specialist achieves 49.3 accuracy, outperforming prior small captioners such as SmallCap (41.8) and Tag2Text (47.1).
When equipped with \MSeR, performance further increases to 50.8, approaching the best generalist pipeline LLaVA-1.5 (51.1) despite using over 14× fewer parameters.
In terms of efficiency, our specialist requires only 4m 53s, and 5m 10s with \MSeR, which is substantially faster than generalist MLLMs ($\approx$ 29m).
These results indicate that our lightweight captioner, combined with \MSeR, offers strong suitability for real-world applications.

\vspace{+.2em}\noindent\textbf{Visual analysis of initial caption utilization in \MSeR.}
A key input to our \MSeR\ framework is the initial caption, which may reflect a rough and first-pass view of the image.
To understand how this caption guides refinement, we examine the model’s visual attention during the generation process.
{\small (\textit{Setup:})}
We analyze the regions the model attends to when generating specific words and visualize attention maps using code adapted from API~\cite{yu2024attention}.
Further implementation details are provided in \secref{sec:added_implementation_attention}.
{\small (\textit{Results:})}
As shown in \figref{fig:attention}, the single-pass captioner often distributes its attention broadly across the image, struggling to localize fine-grained regions.
This behavior reflects the limitation of describing an image in a single glance, where the model attempts to process all information at once without focusing on details.
In contrast, when the model performs the refinement step using the previously generated caption, the attention patterns become more concentrated on the relevant regions associated with each word.
This suggests that the model leverages the initial caption as a guide, enabling it to “look at what matters’’ during refinement.

\begin{table*}[t]
\caption{{\textbf{\MSeR\ on larger language models.}
We evaluate detailed captioning performance on ShareGPT4V~\cite{chen2024sharegpt4v} and DCI~\cite{dci} using captioning specialists built from OPT-1.3B and LLaMA-2-7B.
Across both models, \MSeR\ consistently improves performance over their respective baselines, demonstrating that the refinement framework generalizes beyond lightweight LMs.}
}
\vspace{-.5em}
\tablestyle{2.pt}{1.2}
\resizebox{\linewidth}{!}{%
\begin{tabular}{x{65} x{50} x{40} x{50} ;{.5pt/1.pt} x{50} x{55} ;{.5pt/1.pt} x{50} x{55}}
\shlinesmall
\grgr
 &  & & \multicolumn{1}{c}{}  & \multicolumn{2}{;{.5pt/1.pt} c}{CIDEr~\cite{vedantam2015cider}} & \multicolumn{2}{;{.5pt/1.pt} c}{CAPT~\cite{dong2024benchmarking}} \\[-.1em]
 \grgr
 Language model & vision encoder & LoRA~\cite{hu2022lora} & total \#\,params & initial gen. & + \MSeR & initial gen. & + \MSeR \\
\hdashline
OPT-1.3B~\cite{zhang2022opt} & CLIP ViT-L &  \(\times\) & 1.6B & 50.2 & \textbf{53.1} {\scriptsize (+2.9)} & 49.0 & \textbf{50.2} {\scriptsize (+1.2)} \\
LLaMA-2-7B~\cite{touvron2023llama} & CLIP ViT-L &  \checkmark & 7.3B & 57.3 & \textbf{61.7} {\scriptsize (+4.4)} & 52.7 & \textbf{53.6} {\scriptsize (+0.9)} \\
\shlinesmall
\end{tabular}%
}
\vspace{-1.5em}
\label{tab:larger}
\end{table*}

\vspace{+.2em}\noindent\textbf{\MSeR with larger language models.}
\label{sec:larger_meet}
We extend \MSeR\ to larger language models (LMs), including OPT-1.3B and LLaMA-2-7B, to examine whether the framework generalizes beyond lightweight models.
To build captioning specialists, we first trained these LMs on the ShareGPT and DCI datasets.
We then integrated \MSeR\ and applied an additional refinement stage using the procedure described in \secref{sec:training}.
Our results in \tabref{tab:larger} show that \MSeR\ provides consistent gains over these stronger baselines.
Specifically, the framework improves CAPT~\cite{dong2024benchmarking} scores by 1.2 points for OPT-1.3B and 0.9 points for LLaMA-2-7B, indicating that \MSeR\ remains effective when applied to larger LMs.
Although larger LMs such as OPT-1.3B and LLaMA-2-7B offer stronger captioning capability as observed in prior scaling studies~\cite{kaplan2020scaling, wei2022emergent, hu2022scaling}, their scale, approximately 10× and 56× larger than OPT-125M, can limit practical deployment.
Developing captioning models that balance accuracy and efficiency, therefore, remains an important direction.

\begin{table}[t]
\caption{\textbf{Effect of iterative refinement.} We compare initial captions and one to three refinement steps on the detailed captioning benchmarks ShareGPT4V~\cite{chen2024sharegpt4v} and DCI~\cite{dci}.}
\vspace{-.5em}
\tablestyle{4pt}{1.2}
\resizebox{1.\linewidth}{!}{%
\begin{tabular}{x{47} x{52} x{47} x{47}}
\shlinesmall
\hdashline
\rowcolor[gray]{0.95}
Language model  & Stage & CAPT~\cite{dong2024benchmarking} & GPT~\cite{mllmjudge} \\
\hdashline
\multirow{4}{*}{OPT-125M~\cite{zhang2022opt}}  & initial caption  &  45.9  & 2.74$\pm$0.12 \\
                                               & refinement $\times$1   &  \textbf{48.4}  & \textbf{3.02$\pm$0.12} \\
             & refinement $\times$2 & 47.8 & 3.01$\pm$0.10 \\
             & refinement $\times$3 & 48.1 & 3.03$\pm$0.10 \\
\shlinesmall
\multirow{4}{*}{OPT-1.3B~\cite{zhang2022opt}}    & initial caption  & 49.0     & 3.04$\pm$0.10        \\
                                                 & refinement $\times$1 & 50.2      & 3.14$\pm$0.11         \\
                                                 & refinement $\times$2 & \textbf{50.5}& 3.16$\pm$0.10         \\
                                                 & refinement $\times$3 & 50.3& \textbf{3.20$\pm$0.10}   \\
\shlinesmall
\label{tab:iterative2}
\end{tabular}
}
\vspace{-2.5em}
\end{table}

\vspace{+.2em}\noindent\textbf{Iterative self-refinement.}
\label{sec:iterative}
We also examine whether \MSeR\ benefits from iterative refinement rather than a single refinement pass.
As shown in \tabref{tab:iterative2}, applying multiple refinement steps to the smallest LM, OPT-125M, provides no measurable improvement.
In contrast, two or three refinement iterations yield meaningful gains for OPT-1.3B.
This gap suggests that smaller LMs lack the capacity required to leverage multi-step refinement, whereas larger LMs are able to utilize the additional refinement signal effectively.
The trend aligns with recent findings in LLM-based self-refinement~\cite{selfrefine, ranaldi2024self, wu2024large, kumar2024training}, which report increasing benefits as model capacity grows.
Future work could explore scalable strategies, such as dynamically adjusting the iteration count, which remains a challenge in LLM refinement. 
We note that our core contribution is the demonstration that self-refinement can be effective paradigm for MLLMs.

\vspace{+.2em}\noindent\textbf{Remarks.}
We also evaluate our model on a resource-constrained device, Jetson Nano, in Section~\ref{sec:jetson}. The strategy for generating pseudo-initial captions is detailed in Section~\ref{sec:strategy_hat_c}, and we discuss the limitations of our specialist captioner and existing MLLMs in Section~\ref{sec:discussions}.

\vspace{-.1em}
\section{Related Work}
\vspace{-.3em}

\vspace{+.2em}\noindent\textbf{Multimodal Large Language Models (MLLMs)} have attracted considerable research attention due to their versatile applications, such as chat-bots~\cite{wu2023multimodal}.
Early approaches integrated contrastive image-language pretrained models~\cite{wang2022git} with powerful LLMs, enabling complex reasoning.
The development of instruction-based datasets~\cite{llava} and innovative training strategies~\cite{li2022mplug,ye2024mplug} has further accelerated progress, substantially improving MLLM performance and broadening their capabilities.
Despite these achievements, most MLLMs heavily depend on large-parameter LLMs, making deployment on memory-constrained devices unfeasible. This limitation will likely restrict access for a significant portion of users worldwide.


\vspace{+.2em}\noindent\textbf{Image Captioning Models.}
Recent advances in image captioning have improved training efficiency and descriptive fidelity. 
Approaches such as CaMEL~\cite{barraco2022camel} and SmallCap~\cite{ramos2023smallcap} emphasize minimizing \textit{trainable} parameters by leveraging mean-teacher distillation and employing retrieval augmentation, 
while Tag2Text~\cite{huang2023tag2text} and LoCCa~\cite{wan2024locca} introduce novel mechanisms such as dedicated tagging and location-aware refinement to improve caption quality.
Unlike existing approaches that focus on reducing \textit{trainable} parameters, or rely on single-pass inference—potentially missing crucial details—our method prioritizes \textit{inference} efficiency, considering on-device operation and systematically addressing the limitations of single-pass generation.



\vspace{+.2em}\noindent\textbf{Self-Refinement in LLMs.}
Humans often refine their writing through iterative review to enhance clarity and precision~\cite{selfrefine, dou2024stepcoder}. Recent research has applied this refinement concept to LLMs, introducing techniques. 
For instance, Self-Refine~\cite{le2022coderl, ranaldi2024self, pan2023automatically} enabled models to autonomously critique and iteratively enhance their outputs. 
Unlike such approaches limited to the LLM domain, our method introduces refinement in a multimodal context, guided by both language and vision.


\vspace{-.2em}
\section{Conclusion \& Broader Impact}
\vspace{-.2em}


We introduced a lightweight captioning approach with multimodal self-refinement, motivated by the need for efficient visual understanding systems on edge devices.
Our study began with an OPT-125M–based captioner, which surprisingly achieved performance comparable to large MLLMs despite a substantial reduction in parameters (93\%{\footnotesize$\downarrow$}) and inference time (97\%{\footnotesize$\downarrow$}).
To further enhance this lightweight model, we proposed a multimodal self-refinement framework, the first refinement-based approach explored in the MLLM community.
Through extensive experiments, we demonstrated that the proposed architecture and framework provide more accurate and informative captions.
These improvements hold consistently across single-sentence captioning, detailed captioning, and practical downstream tasks such as long-range video question answering.
We hope that the model explored in this work serves as a practical solution for on-device applications and that our findings on visual self-refinement inspire deeper investigation in future multimodal research.

\vspace{+.5em}\noindent\textbf{Broader Impact.} 
Future research may explore integrating external tools (e.g., zoom or crop, as in GPT-o3~\cite{openai2024o3o4}), and designing a unified multimodal connector for both the initial and secondary glance. More research questions that aim to enhance both captioning performance and efficiency in real-world applications are provided in \secref{sec:questions}.



{
    \small
    \bibliographystyle{ieeenat_fullname}
    \bibliography{main}
}

\newpage
\renewcommand \thepart{}
\renewcommand \partname{}

\appendix

\clearpage
\vspace{-5em}


{\LARGE 
    \noindent\textbf{Appendix}
}

\vspace{-1em}
\section{Novelty of our framework \MSeR}
\label{sec:summary}




\begin{itemize}[leftmargin=15pt, itemsep=1pt]
    \item \textbf{What is it?} Humans first take in the overall scene, then refine at specific regions to notice finer details. Our \MSeR\ framework mimics this human tendency, allowing the captioning specialist to revise its output.
    \item \textbf{Why novel?} To the best of our knowledge, this is the first work to introduce a multimodal refinement method that jointly utilizes visual features and the model's output.
\end{itemize}

\begin{itemize}[label=\(\circ\), leftmargin=15pt, itemsep=1pt]
        \item \textit{Lightweight captioning community:} \textbf{This field has been gradually declining since the remarkable capabilities of LLMs were discovered. In this work, we revisit the practically important yet underexplored topic of lightweight captioning models.}
        \begin{itemize}[label=\(*\), leftmargin=15pt, itemsep=1pt]
            \item Unlike ours, some methods~\cite{luo2023semantic, ramos2023smallcap, wang2022text} utilize prior captions obtained through a heavy image-text retrieval process. For example, SmallCap performs similarity matching between a given image and 500,000 candidate captions at each iteration.
            \item Unlike ours, several works~\cite{huang2023tag2text, lu2018neural} incorporate object detection or tagging procedures. However, despite these extra components, their captioning performance has been limited.
            \item Unlike ours, which refines text output generated via a full forward pass, certain works~\cite{luo2023semantic, chen2022analog} adopt Diffusion Transformers and denoise latent text embeddings. Moreover, they provide little motivation or analysis as to why such a process is important from the perspective of utilizing visual cues more effectively.
        \end{itemize}
        \item \textit{NLP community}: Humans often refine their writing, and coders revise their code through iterative review. Recent research has applied this refinement/correctness concept to LLMs~\cite{selfrefine, ranaldi2024self, wu2024large, kumar2024training}. Unlike such approaches limited to the LLM domain, our method introduces refinement in a multimodal context.
        \item \textit{Multimodal LLMs community:} Existing models~\cite{llava, bai2023qwenvl, zhu2025internvl3} heavily rely on the complex reasoning capabilities of LLMs and typically generate outputs in a single pass. We uniquely demonstrate the effectiveness of revisiting and refining its own textual outputs.
        \item \textit{Visual Blindness community:} Several studies~\cite{cambrian, eyeswideshut} believe the visual encoder is a critical bottleneck, necessitating adapting multiple vision encoders. Since our research focuses on a lightweight model, we maximize the utility of the existing vision encoder through multi-level utilization.
    \end{itemize}



\section{Additional Experiments}

\subsection{When GPT meets `self-refinement'}
\label{sec:gpt_meet}
\begin{figure*}[t]\centering
\includegraphics[width=1.\linewidth]{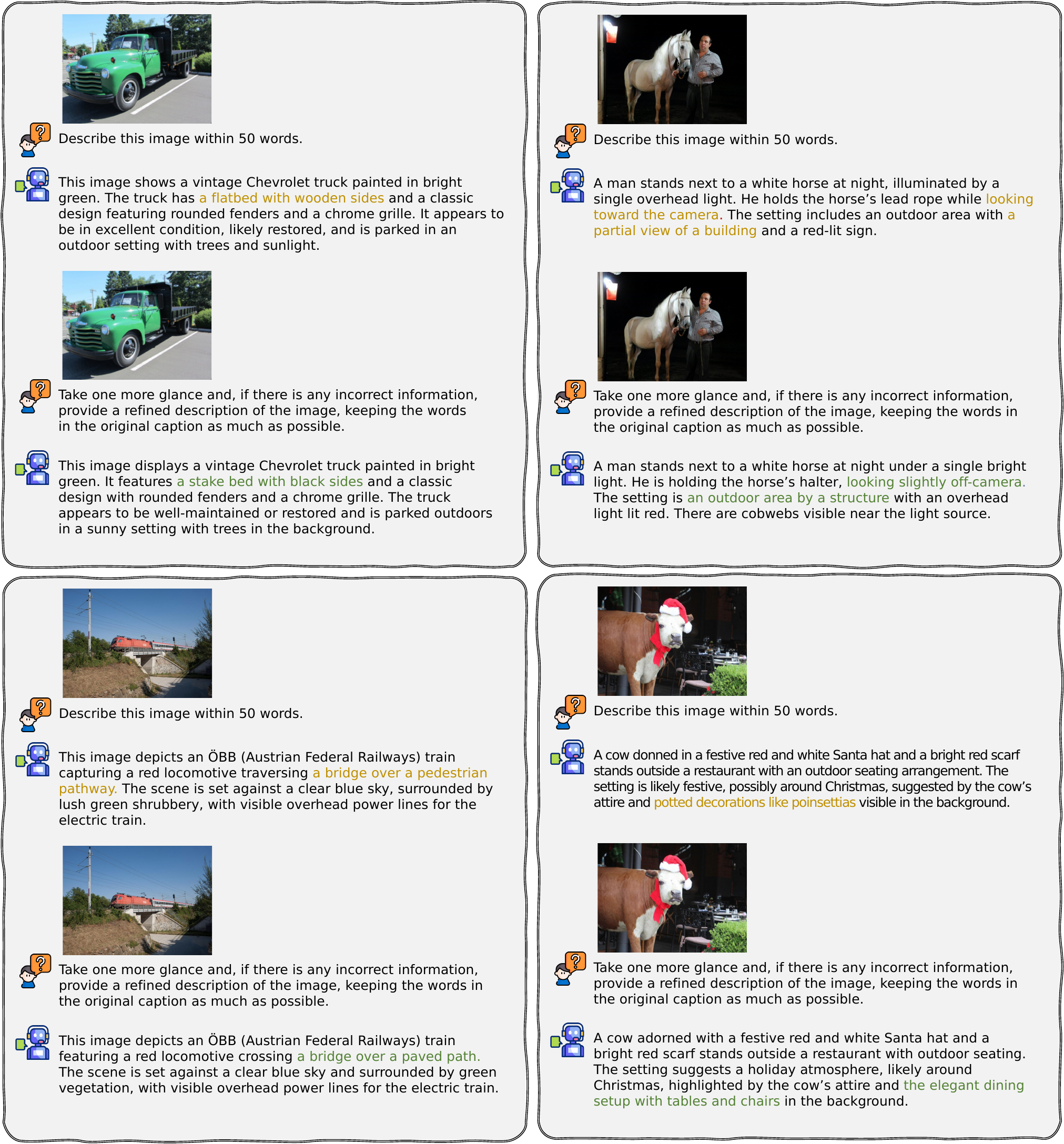}
\vspace{-1.em}
\caption{\textbf{When GPT meets self-refinement.} Example outputs from OpenAI's GPT before and after being prompted to self-refinement.}
\label{fig:gpt}
\end{figure*}

We additionally examine how OpenAI’s GPT models, which show strong captioning performance, behave under this refinement process.
Although our framework includes two components, an previous caption and visual representations from SeR-connector, we apply only the former, since modifying the internal architecture of GPT models is not feasible. The results, shown in \figref{fig:gpt}, are obtained using GPT-4-turbo.
As a commercial model with a large parameter budget, GPT-4-turbo often generates accurate initial captions, though occasional inaccuracies still appear. To assess whether refinement is beneficial, we instruct the model to perform \textbf{self-refinement} and observe that it can revise its caption by re-examining the provided image.
OpenAI’s GPT-o3 reflects similar ideas through the introduction of Thinking with Images~\cite{openai2024o3o4}. However, this direction remains relatively unexplored in the research community, indicating the need for further investigation.

\subsection{Evaluation on an Actual Edge Device}
\label{sec:jetson}

o highlight the practical value of our approach, we evaluate our lightweight captioner on edge devices, where deploying large-scale MLLMs is often infeasible. Our motivation stems from the observation that, although MLLMs are powerful, their computational demands limit use in resource-constrained environments. In contrast, lightweight captioners remain relatively unexplored despite their suitability for real-world applications.
We show that such captioners can be effectively deployed on devices including an RTX 3090 and the \textbf{Jetson Nano}. We assess captioning performance on the MS COCO and ShareGPT4V\&DCI datasets, and additionally measure inference time, memory usage, and power consumption. All models are executed for 100 iterations with a batch size of 1. As shown in \tabref{tab:resource_eval}, our method performs consistently across devices and, importantly, remains fully operational in settings where models like LLaVA-1.5 cannot run. These results support the deployability of our framework on edge hardware and highlight lightweight captioning as a promising direction for real-world assistive technologies.

\subsection{Comparison on different learning strategies}
\label{sec:dpo}

We examine how our lightweight specialist performs under different learning strategies.
The first strategy trains the model with maximal data coverage by combining COCO, ShareGPT, DCI, and GLaMM (\ie, more data).
The second strategy applies distillation, training on captions generated by LLaVA-1.6-34B~\cite{llava-recap} (\ie, distillation).
Our original approach trains the model solely on the target datasets (\ie, origin).
The results in \tabref{tab:moredata} show that the more data strategy yields limited improvement, likely due to weakened task alignment when mixing heterogeneous datasets. In contrast, the distillation strategy benefits from learning from a strong teacher and improves performance even in a single-pass setting. When combined with our refinement framework (\ie, \MSeR), it produces further gains, suggesting that distillation and \MSeR complement each other.
Additionally, as noted in \secref{sec:training}, our datasets resemble those used in Direct Preference Optimization (DPO). Exploring reinforcement-learning-based training such as DPO may offer another promising direction for improving lightweight captioners.

\begin{table*}[t]
\tablestyle{4pt}{1.2}
\caption{Evaluation results across different hardware resources and datasets.} 
\vspace{-1em}
\resizebox{1.\linewidth}{!}{%
\begin{tabular}{x{50} x{40} x{50} x{25} x{25} x{25} x{25} x{25} x{30} x{30}}
\shlinesmall
\hdashline
\rowcolor[gray]{0.95}
Resource (RAM) & Data & Model & Mem. & Inf. time & Power. & B@4~\cite{papineni2002bleu} & CIDEr~\cite{vedantam2015cider} & CLAIR~\cite{clair} & GPT~\cite{mllmjudge} \\
\hdashline
Jetson\,Nano\,(4G) & All & LLaVA-1.5-7B & out-of-memory & N/A & N/A & N/A & N/A & N/A & N/A \\
\hdashline
RTX 3090 (24G) & \multirow{2}{*}{MS COCO~\cite{mscoco}} & \multirow{2}{*}{Ours-500M} & 3.2G & 5s  & 230 W & 39.5 & 133.8 & 78.6$\pm$2.9 & 2.83$\pm$0.06 \\
Jetson\,Nano\,(4G)& & & 2.6G & 20s & 13 W  & 39.5 & 133.8 & 78.6$\pm$2.9 & 2.73$\pm$0.09 \\
\hdashline
RTX 3090 (24G) & ShareGPT4V~\cite{chen2024sharegpt4v} & \multirow{2}{*}{Ours-500M} & 3.2G & 5s  & 230 W & 22   & 43.2 & 57.9$\pm$3.0 & 3.01$\pm$0.10 \\
Jetson\,Nano\,(4G)& \& DCI~\cite{dci} & & 2.7G & 21s & 13 W  & 22.2 & 42.9 & 57.4$\pm$2.9 & 3.02$\pm$0.11 \\
\shlinesmall
\end{tabular}
}
\label{tab:resource_eval}
\end{table*}

\begin{table*}[t]
\tablestyle{2.pt}{1.2}
\caption{Performance comparison of our lightweight captioner under different learning strategies: origin, more data, and distillation. Distillation from a strong teacher, especially when combined with \MSeR, leads to the best results.}
\vspace{-1em}
\newcolumntype{g}{>{\columncolor[gray]{0.95}} x{50}}
\newcolumntype{k}{>{\columncolor[gray]{0.95}} x{40}}
\resizebox{1.\linewidth}{!}{%
\begin{tabular}{x{65} ;{1.5pt/.5pt} g g k ;{1.5pt/.5pt} x{55} ;{1.5pt/.5pt} x{50} x{50} x{40} } 
\shlinesmall
\grgr
\multicolumn{8}{c}{ShareGPT4V~\cite{chen2024sharegpt4v} \& DCI~\cite{dci}} \\
\hdashline
Metric & \textbf{origin} & +SeR & gain & \textbf{more data} & \textbf{distillation} & +SeR & gain \\
\hdashline
CIDEr~\cite{vedantam2015cider} & 40.5 & 43.6 & {\scriptsize +3.1} & 36.8  & 42.6 & 43.6  & {\scriptsize +2.0} \\
CAPT~\cite{dong2024benchmarking}  & 45.9 & 48.4 & {\scriptsize +2.5} & 43.2  & 46.5 & 47.4  & {\scriptsize +0.9} \\
\shlinesmall
\end{tabular}
}
\vspace{-1em}
\label{tab:moredata}
\end{table*}

\subsection{Efficacy with Other Vision Encoders}
\label{sec:other_vision}

We examine whether our strategy of selecting multi-level features generalizes to vision encoders beyond CLIP~\cite{CLIP}, which serves as our original setup. As shown in \tabref{tab:vision_encoder_comparison}, the approach consistently improves performance across different encoders. Both SigLIPv2~\cite{siglip2} and DINOv2~\cite{oquab2024dinov} benefit from incorporating multi-level features, indicating that our method is not restricted to CLIP-based models.
Interestingly, CLIP with multi-level features surpasses the combination of CLIP and DINOv2 in several metrics while maintaining better parameter efficiency. In contrast, DINOv2 alone delivers lower performance, likely due to weaker alignment with language features. For all experiments, we pair OPT-125M with each vision encoder and train the resulting models on the ShareGPT~\cite{chen2024sharegpt4v}\&DCI~\cite{dci} datasets.

\begin{table*}[t]
\caption{For the captioning specialist, we evaluate different vision encoders and the corresponding selected layer indices.}
\vspace{-1em}
\tablestyle{4pt}{1.2}
\resizebox{1.\linewidth}{!}{%
\begin{tabular}{x{85} x{40} x{75} x{52} x{52} x{52}}
\shlinesmall
\hdashline
\rowcolor[gray]{0.95}
Vision encoder & \#params & indices of selected layers & CIDEr~\cite{vedantam2015cider} & CLAIR~\cite{clair} & GPT~\cite{mllmjudge} \\
\hdashline
CLIP~\cite{CLIP} & 300M & \{23\}         & 42.8 & 55.8 & 2.78 \\
            & 300M & \{13, 18, 23\} & 43.3 & 57.7 & 3.02 \\
\hdashline            
CLIP~\cite{CLIP}+DINOv2~\cite{oquab2024dinov} & 600M & \{23\} + \{23\} & 42.9 & 57.3 & 3.02 \\
\hdashline
SigLIPv2~\cite{siglip2}    & 300M & \{23\}         & 43.0 & 56.2 & 2.88 \\
            & 300M & \{15, 23\}     & \textbf{45.9} & 57.7 & 3.05 \\
            & 300M & \{13, 18, 23\} & 45.5 & \textbf{58.2} & \textbf{3.07} \\
\hdashline
DINOv2~\cite{oquab2024dinov}      & 300M & \{23\}         & 32.8 & 48.6 & 2.55 \\
            & 300M & \{13, 18, 23\} & 33.0 & 50.1 & 2.66 \\
\shlinesmall
\end{tabular}
}
\vspace{-1.5em}
\label{tab:vision_encoder_comparison}
\end{table*}




\section{Additional Related Work}
\label{sec:add_related}

\subsection{Visual Blindness in VLMs.}
Despite significant advancements, MLLMs still face limitations in their visual capabilities, hindering their practical applications. 
Eyes Wide Shut~\cite{eyeswideshut} demonstrated that even GPT-4V~\cite{achiam2023gpt} struggles with basic visual questions. 
Research on this topic typically points to two main sources of failure: one relates to the language decoder, which can hallucinate details not present in the image~\cite{bai2024hallucination}, while the other focuses on the visual encoder, which may provide ambiguous visual information. 
Several studies, including Cambrian~\cite{cambrian}, suggest that the visual encoder provides ambiguous visual information and constitutes a critical bottleneck.
We also concentrate on the visual issue, particularly within the context of our lightweight model, where the visual encoder accounts for a significant portion of the parameters. Furthermore, we introduce a novel operational framework to improve visual grounding.

\section{SeR-Connector}
\label{sec:deeplens_detail}

\begin{figure}[t]\centering
\includegraphics[width=.9\linewidth]{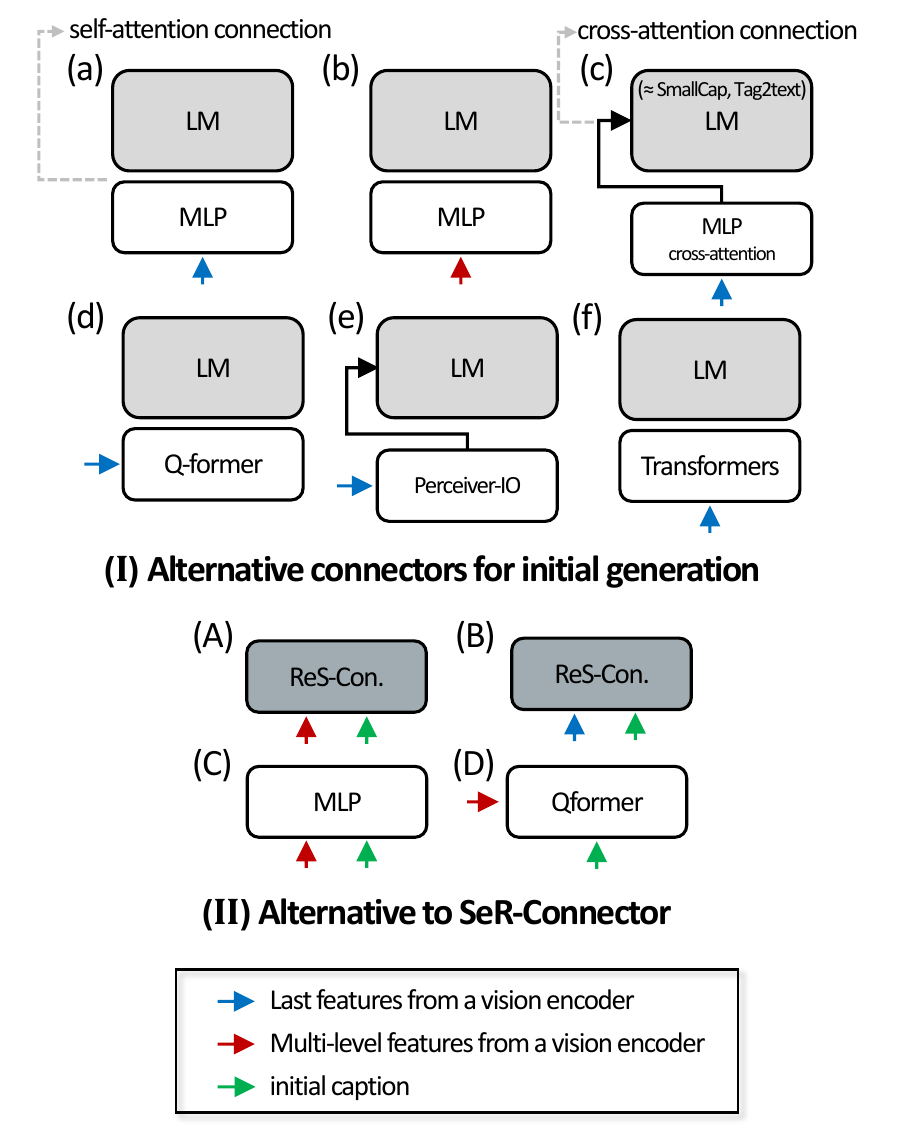}
\vspace{-.5em}
\caption{Ablation study of multimodal connector designs. Each configuration corresponds to evaluation in \tabref{tab:ablation}.}
\vspace{-.5em}
\label{fig:deeplens}
\end{figure}


\subsection{Ablation study}
\label{sec:deeplens_ablation}
We conduct a series of ablation experiments to analyze the design choices behind the SeR-Connector, as summarized in \tabref{tab:ablation}, with the corresponding variants illustrated in \figref{fig:deeplens}. Our analysis covers three aspects: connector selection, refinement configurations, and the layer-indexing strategy for visual feature extraction. \textbf{(I)} We first compare multimodal connectors used for initial caption generation. Prior work has explored designs such as Cross-Attention~\cite{ramos2023smallcap, barraco2022camel}, Q-Former~\cite{li2023blip2bootstrappinglanguageimagepretraining}, and Transformer-style modules, yet we find that the simple MLP connector from LLaVA~\cite{llava15} remains competitive. Configurations (b) and (f), which incorporate multi-level features and a BERT-style Transformer block, respectively, show slightly improved performance, but given the small gains, we preserve the lightweight MLP structure for efficiency. \textbf{(II)} We then assess different connectors for the refinement stage. Combining our base configuration (a) with structure (A), which uses both proposed inputs, achieves strong results. Although applying (A) to configuration (b) yields a minor improvement, it is insufficient to justify adopting it as the default. \textbf{(III)} Lastly, we examine the effect of selecting different layer sets from the ViT encoder. Across the tested combinations, using a diverse trio of layers {13, 18, 23} provides the best performance.

\begin{table*}[t]
\vspace{1em}
\caption{Token-level differences between pseudo-initial and ground-truth captions.}
\tablestyle{4pt}{1.2}
\vspace{-1em}
\resizebox{0.95\linewidth}{!}{%
\begin{tabular}{x{180} ;{.5pt/1.pt} x{190}}
\shlinesmall
\hdashline
\rowcolor[gray]{0.95}
pseudo-initial caption $\hat{c}$ & different tokens $E ={t \mid \hat{c}(t) \neq c(t)}$ in \secref{sec:training} \\
\hdashline
A woman in a room with two dogs 
  & two / dogs \\
A cat sitting on a chair in front of the window. 
  & sitting / on a chair / in front of the window \\
\shlinesmall
\end{tabular}
}
\label{tab:examples}
\end{table*}


\begin{table}[t]
\caption{Ablation results evaluating connector types, refinement configurations, and ViT layer selections in SeR-Connector. Architectural variants are illustrated in \figref{fig:deeplens} (b) and (c).}
\vspace{-.5em}
  \centering
  \begin{minipage}{\linewidth}
    \centering
    \resizebox{\linewidth}{!}{%
      \begin{tabular}{x{40} x{40} x{40} x{40} x{40} x{40}}
        \shlinesmall
        \grgr
        \multicolumn{6}{c}{\,\textbf{(I)} connectors for initial generation} \\[+0.1em]
        \colorbox{highlight!75}{\makebox(6,6){\strut\textcolor{black}{(a)}}} & (b) & (c) & (d) & (e) & (f)\\
        \hdashline
        129.6 \checkmark & 130.6 & 125.9 & 127.7 & 122.9 & 130.9 \\
        \shlinesmall
      \end{tabular}%
    }
  \end{minipage}
  \\[+.5em]
  \begin{minipage}{\linewidth}
    \centering
    \resizebox{\linewidth}{!}{%
      \begin{tabular}{x{50} x{50} x{50} x{50} x{70}}
        \shlinesmall
        \grgr
        & \multicolumn{3}{c}{\,\textbf{(II)} \colorbox{highlight!75}{\makebox(6,6){\strut\textcolor{black}{(a)}}} + connectors for SeR} 
          & \multicolumn{1}{;{0.2pt/1.5pt}c}{(b) + con.} \\[+0.1em]
        (A) & (B) & (C) & (D) & \multicolumn{1}{;{0.2pt/1.5pt}c}{(A)} \\
        \hdashline
        133.5 \checkmark & 131.9 & 132.1 & 131.9 & \multicolumn{1}{;{0.2pt/1.5pt}c}{133.6} \\
        \shlinesmall
      \end{tabular}%
    }
  \end{minipage}
  \\[+.5em]
  \begin{minipage}{\linewidth}
    \centering
    \resizebox{\linewidth}{!}{%
      \begin{tabular}{x{50} x{50} x{45} x{50} x{50}}
        \shlinesmall
        \grgr
        \multicolumn{5}{c}{\,\textbf{(III)} indexes of selected layers in ViT} \\[+0.1em]
        \{23\} & \{13, 23\} & \{15, 23\} & \{15, 19, 23\} & \{13, 18, 23\} \\
        \hdashline
        131.9 & 133.0 & 133.0 & 133.3 & 133.5 \checkmark \\
        \shlinesmall
      \end{tabular}%
    }
  \end{minipage}
  \vspace{-.5em}
  \label{tab:ablation}
\end{table}

\section{Discussions}
\label{sec:discussions}

\subsection{Why do previous small models fall short?}
\label{sec:whyexisting}
The development of captioning models can be viewed in two phases: before and after the integration of LLMs. Earlier approaches typically relied on architectures with relatively small parameter counts. For instance, CaMEL~\cite{barraco2022camel} and SmallCap~\cite{ramos2023smallcap} used GPT-2 models with 125M to 350M parameters, while Tag2Text~\cite{huang2023tag2text} and LoCCa~\cite{wan2024locca} employed BERT-based models ranging from 300M to 900M parameters. The emergence of ClipCap~\cite{mokady2021clipcap}, BLIP~\cite{li2023blip2}, and LLaVA~\cite{llava} shifted the field toward LLM-driven captioners, and recent research has largely centered on building MLLMs.
In contrast to this trend, we revisit smaller captioning models and highlight an overlooked limitation. Many of these models inject visual features through cross-attention, a design that offers limited benefit when paired with small language models. Empirically, this is reflected in the performance of structure (c) in \figref{fig:deeplens}, which yields a relatively low score of 125.9 in \tabref{tab:ablation}. Had the field not shifted so strongly toward LLMs, such architectural constraints in small models might have been recognized earlier.
Building on this insight, we adopt the LLaVA architecture and inject visual features directly into the self-attention inputs, as shown in \figref{fig:deeplens} (a). This simple modification leads to stronger performance and demonstrates that small models can remain practical and effective. \textbf{We hope this encourages reducing reliance on LLMs for tasks such as captioning and fosters greater interest in developing lightweight yet capable models.}

\subsection{Limitations of Existing MLLMs}
\label{sec:whyexisting_mllm}

\begin{figure*}[t]\centering
\includegraphics[width=1.0\linewidth]{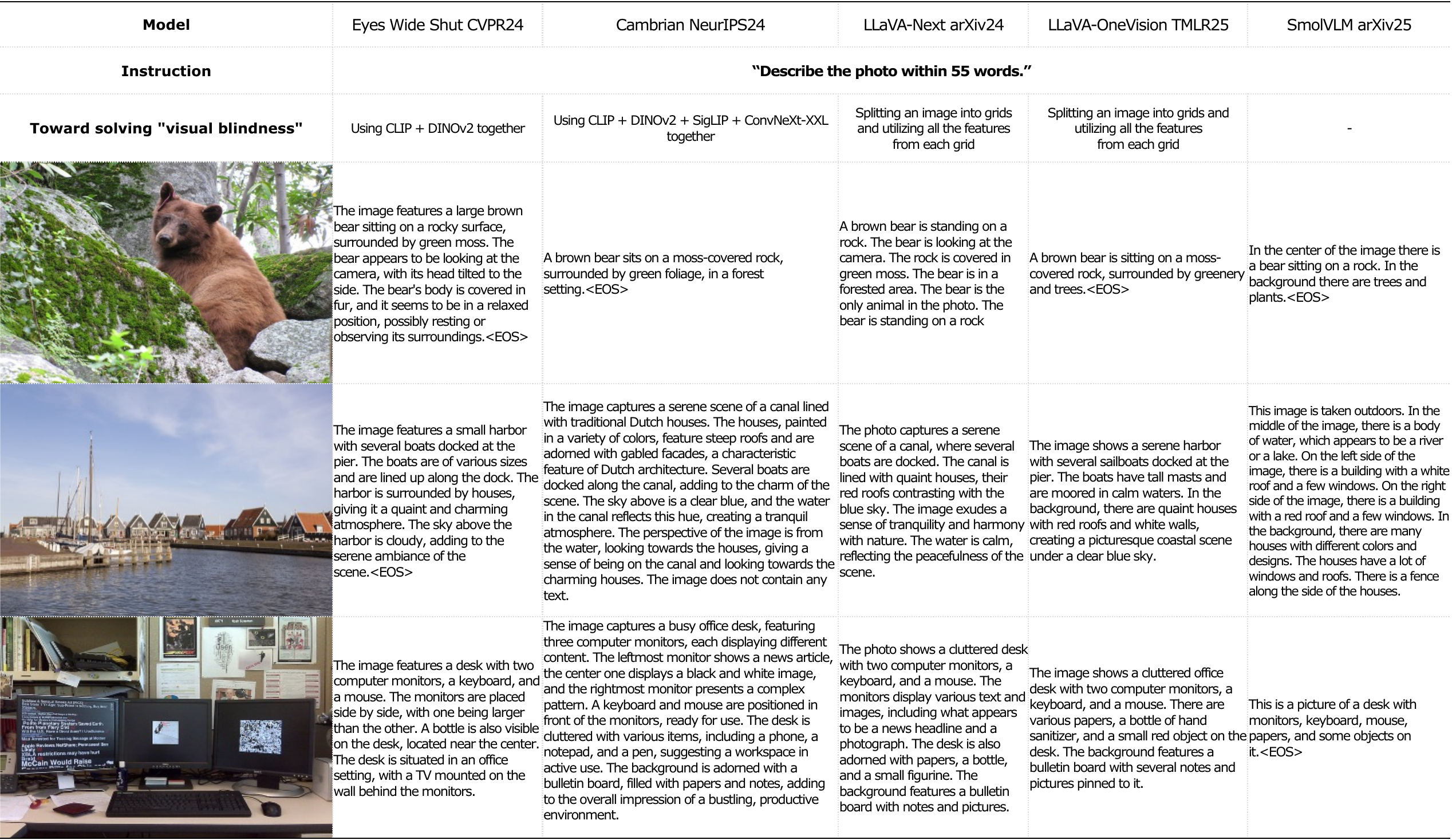}
\vspace{-1.5em}
\caption{Despite their size, existing MLLMs struggle with visual blindness, especially in complex scenes. We hope our approach offers a meaningful step toward alleviating this issue.}
\vspace{-1em}
\label{fig:limitation_mllm}
\end{figure*}

We examine the broader challenges faced by existing multimodal large language models (MLLMs), particularly their susceptibility to visual blindness. Prior work, including Eyes Wide Shut~\cite{eyeswideshut} and Cambrian~\cite{cambrian}, has identified this issue and attempted to mitigate it using multiple vision encoders such as DINOv2~\cite{oquab2024dinov}, SigLIPv2~\cite{siglip2}, and CLIP~\cite{CLIP}. However, as illustrated in \figref{fig:limitation_mllm}, even large-scale models continue to struggle with producing consistent long-form captions in complex, multi-object scenes. We further evaluate two recent MLLMs, LLaVA-Next~\cite{llavanext} and LLaVA-OneVision~\cite{llavaonevision}. Despite employing advanced techniques—such as partitioning the input into grids and processing features from each region—both models still generate incorrect captions.
These observations indicate that visual blindness remains a persistent issue across different model sizes and architectures. In this context, our \MSeR\ framework, which directs attention to key regions via initial captions and leverages multi-level features from a single vision encoder, offers an efficient and effective step toward addressing this limitation.

\subsection{Limitations of lightweight captioners}
\label{sec:limitation}
\begin{figure*}[t]\centering
\includegraphics[width=1.0\linewidth]{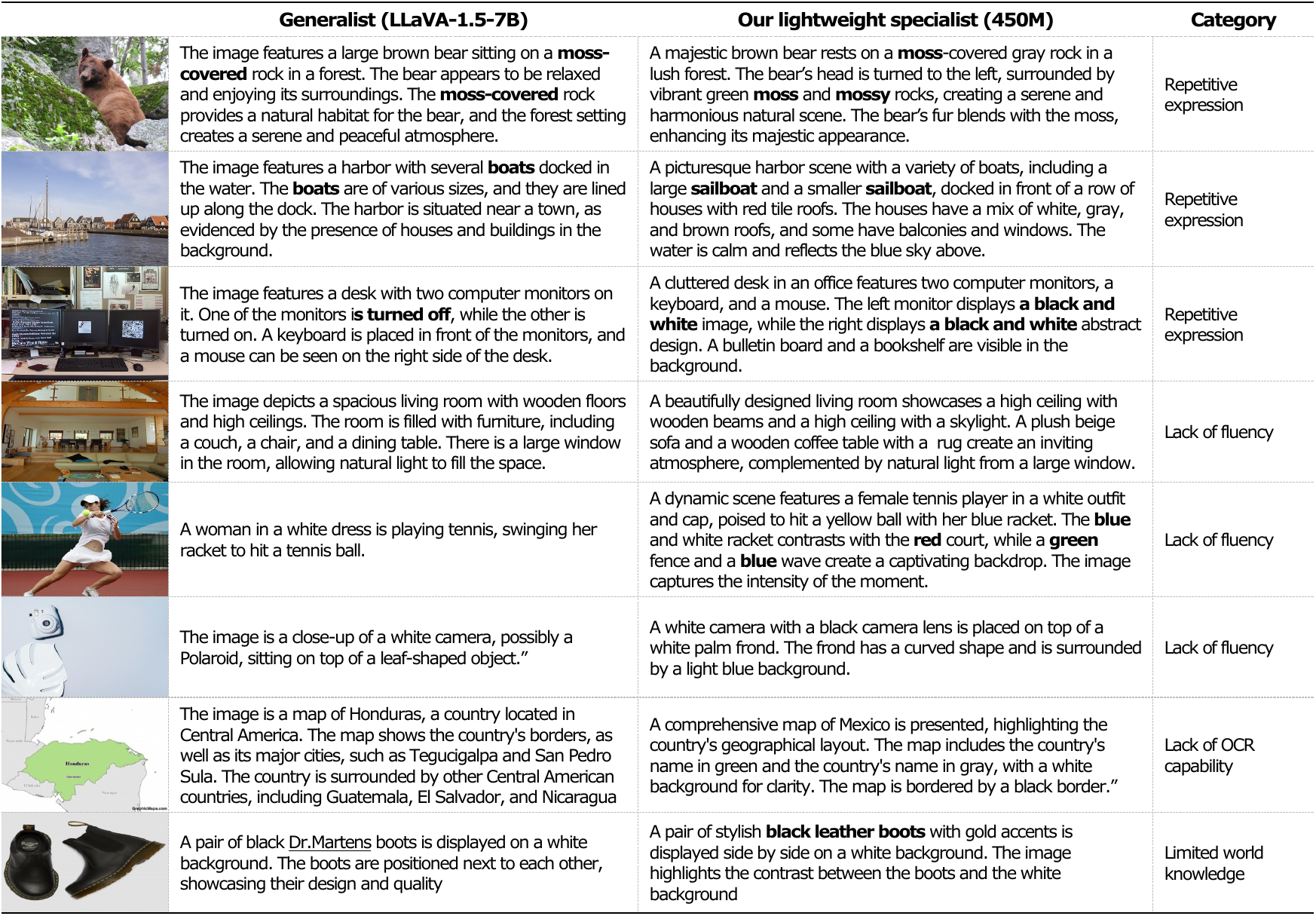}
\vspace{-1.5em}
\caption{Despite their efficiency, small models suffer from various limitations. Further research is required to assess how high-quality data can compensate for their weaknesses.}
\label{fig:limitation}
\end{figure*}

While we have demonstrated the potential of small models in captioning tasks, the use of lightweight LMs unavoidably introduces some limitations. In particular, we observe that the model occasionally suffers from issues such as repetitive phrasing, reduced fluency, limited OCR capability, and a lack of general world knowledge. Examples of these cases are provided in \figref{fig:limitation}.
These limitations may stem from two primary factors: (\textit{i}) the small number of parameters, which can restrict the model’s capacity for complex reasoning and language generation~\cite{kaplan2020scaling}, and (\textit{ii}) the limited scale and quality of training data, as our model was trained on approximately 500K image-caption pairs from ShareGPT-4V~\cite{chen2024sharegpt4v} which contains machine-generated captions. A natural direction for future work is to investigate how far the capabilities of small models can be scaled with access to \textit{larger and higher-quality training datasets}.
In addition to the results in \secref{sec:whyexisting_mllm}, we observe similar issues in larger models such as LLaVA-1.5, suggesting that these challenges remain unresolved~\cite{bai2024hallucination,eyeswideshut} and require deeper investigation.

\subsection{Limitations of existing evaluation methods}
\label{sec:limitation_eval}
To ensure fair comparison across captioning models, we adopt seven evaluation metrics: BLEU@4~\cite{papineni2002bleu}, METEOR~\cite{meteor}, CIDEr~\cite{vedantam2015cider}, BERTScore~\cite{bertscore}, CAPTURE~\cite{dong2024benchmarking}, CLAIR~\cite{clair}, and MLLM-as-judge~\cite{mllmjudge}. For CLAIR and MLLM-as-judge, we randomly sample 100 images and evaluate each with 10 different seeds to report both the mean and standard deviation (shown with the $\pm$ symbol).
Despite these efforts, current evaluation metrics do not always correlate well with human judgment. Some models rank higher under one metric but lower under others, leading to inconsistent comparisons. Moreover, as discussed in \secref{sec:limitation}, small specialists sometimes exhibit reduced fluency, which existing metrics fail to capture. Among the metrics examined, MLLM-as-judge generally provides more stable and reliable assessments, while CLAIR shows higher variance across runs.
These observations highlight the need for more robust evaluation methods for captioning. Future research should account for multiple aspects of quality, including fluency, coherence, faithfulness, relevance, informativeness, and completeness. Additionally, moving beyond n-gram matching, MLLM-based evaluators (e.g., OpenAI GPT) will be essential for producing consistent and trustworthy assessments. Such advancements can improve confidence in captioning-based applications.



\subsection{Role of Pseudo-Initial Captions in Refinement}
\label{sec:peudo_caption}

In this part, we discuss how incorporating pseudo-initial captions provides more effective supervision during training. We demonstrate the following points in the main paper: 
(\textit{i}) The model is trained to generate the ground-truth (GT) caption given both the image and the pseudo-initial caption as input: $I + \hat{c} \;\rightarrow\; \text{Our model} \;\rightarrow\; c$. 
(\textit{ii}) If the pseudo-initial caption is generated following the strategies in \secref{sec:training}, then it is unlikely to include too many differing tokens from the GT caption.
To illustrate the effect of pseudo-initial captions, we provide an example from \tabref{tab:examples}: consider the pair in which both $c$ and $\hat{c}$ correspond to variations of “A woman in a room with a cat.” When the pseudo-initial caption and the GT caption share a substantial portion of tokens, the model is encouraged to consult $\hat{c}$ and revise only the mismatches to recover the GT caption. In contrast, if the two captions differ too greatly, the model tends to ignore the pseudo-initial caption and regenerate the GT caption independently. This behavior enables the model to correct localized errors rather than rewriting the entire caption, thereby preventing misleading supervision.

\textbf{As an additional experiment,} we fine-tune our captioner in Stage~2 of \figref{fig:workflow} using pseudo-initial captions generated under four conditions: (Data,1) initial captions produced by the Stage~1–trained captioner; (Data,2) pseudo-initial captions that differ substantially from the GT caption; (Data,3) pseudo-initial captions with minor modifications from the GT caption (our default strategy); and (Data,4) two pseudo-initial captions per sample—one generated as in Data,3 and one identical to the GT caption (e.g., GT: “A woman in a room with a cat”; pseudo-initials: “A boy in a room with a dog” and “A woman in a room with a cat”).
\tabref{tab:detailed_captioning} shows that Data,1 yields reasonable improvements within our framework, while Data,2 produces the expected behavior in which the model disregards the pseudo-initial caption and regenerates a new one. Notably, Data,4 performs comparably to Data,3, and together with the effect observed in Data,1, suggests that our framework is \textit{robust} to moderate variation in pseudo-initial caption quality.

\begin{table*}[t]
\tablestyle{4pt}{1.2}
\caption{Detailed captioning results on ShareGPT4V~\cite{chen2024sharegpt4v} \& DCI~\cite{dci} with fine-tuning on four types of pseudo-initial captions. The comparison highlights how different levels of overlap or inclusion of GT captions affect model performance.}
\vspace{-1em}
\resizebox{.95\linewidth}{!}{%
\begin{tabular}{x{115} ;{.5pt/1.pt} x{43} x{30} ;{.5pt/1.pt} x{43} x{30} ;{.5pt/1.pt} x{43} x{30}}
\shlinesmall
\hdashline
\rowcolor[gray]{0.95}
Detailed captioning & CIDEr~\cite{vedantam2015cider} & gain & CAPT~\cite{dong2024benchmarking} & gain & GPT~\cite{mllmjudge} & gain \\
\hdashline
Our specialist fine-tuned via Stage 1 
  & 40.5 & -   & 45.9 & -   & 2.74$\pm$0.12 & - \\
finetuned w/Data1 
  & 42.5 & +2.0 & 47.2 & +1.3 & 2.86$\pm$0.11 & +0.12 \\
finetuned w/Data2 
  & 41   & +0.5 & 46.6 & +0.7 & 2.76$\pm$0.12 & +0.02 \\
finetuned w/Data3 
  & 43.6 & +3.1 & 48.4 & +2.5 & 3.02$\pm$0.12 & +0.28 \\
finetuned w/Data4 
  & 43.8 & +3.3 & 48.2 & +2.3 & 3.04$\pm$0.09 & +0.30 \\
\shlinesmall
\end{tabular}
}
\vspace{-.8em}
\vspace{-.5em}
\label{tab:detailed_captioning}
\end{table*}

\subsection{Further Research Questions}
\label{sec:questions}
To guide future exploration, we outline several research directions that may advance the field of lightweight captioning and multimodal learning more broadly:

\begin{enumerate}[leftmargin=15pt, itemsep=1pt]
\item What is the minimal model size required for a captioning specialist to be practically useful in real-world assistive technologies? At what point does the performance–efficiency trade-off stabilize?

\item Although supervised training is commonly used in multimodal training, recent work has shown clear benefits from reinforcement learning in LLMs~\cite{touvron2023llama}. Can similar reward-driven methods improve captioning models?

\item The LLaVA framework is widely adopted due to its simplicity and open-source accessibility. Could a unified Transformer-based multimodal architecture serve as a more effective foundation for practical captioners?

\item Multimodal performance depends heavily on the quality of visual representations. How reliable are features from encoders such as SigLIP~\cite{zhai2023sigmoid} or MAE~\cite{mae_survey}, especially for fine-grained captioning?

\item Can our framework be applied effectively in simpler VQA settings, and are large language models necessary in such cases?

\item Do captioning models exhibit biases (e.g., gender, occupation, age), and how should such concerns be addressed across diverse cultural contexts and deployment domains such as indoor, outdoor, and robotic environments? Could domain adaptation with user feedback offer a viable solution?

\end{enumerate}

We consider these questions promising directions for future research and believe that addressing them will contribute to more widely deployable multimodal captioning systems.


\begin{figure*}[t]\centering
\includegraphics[width=1.\linewidth]{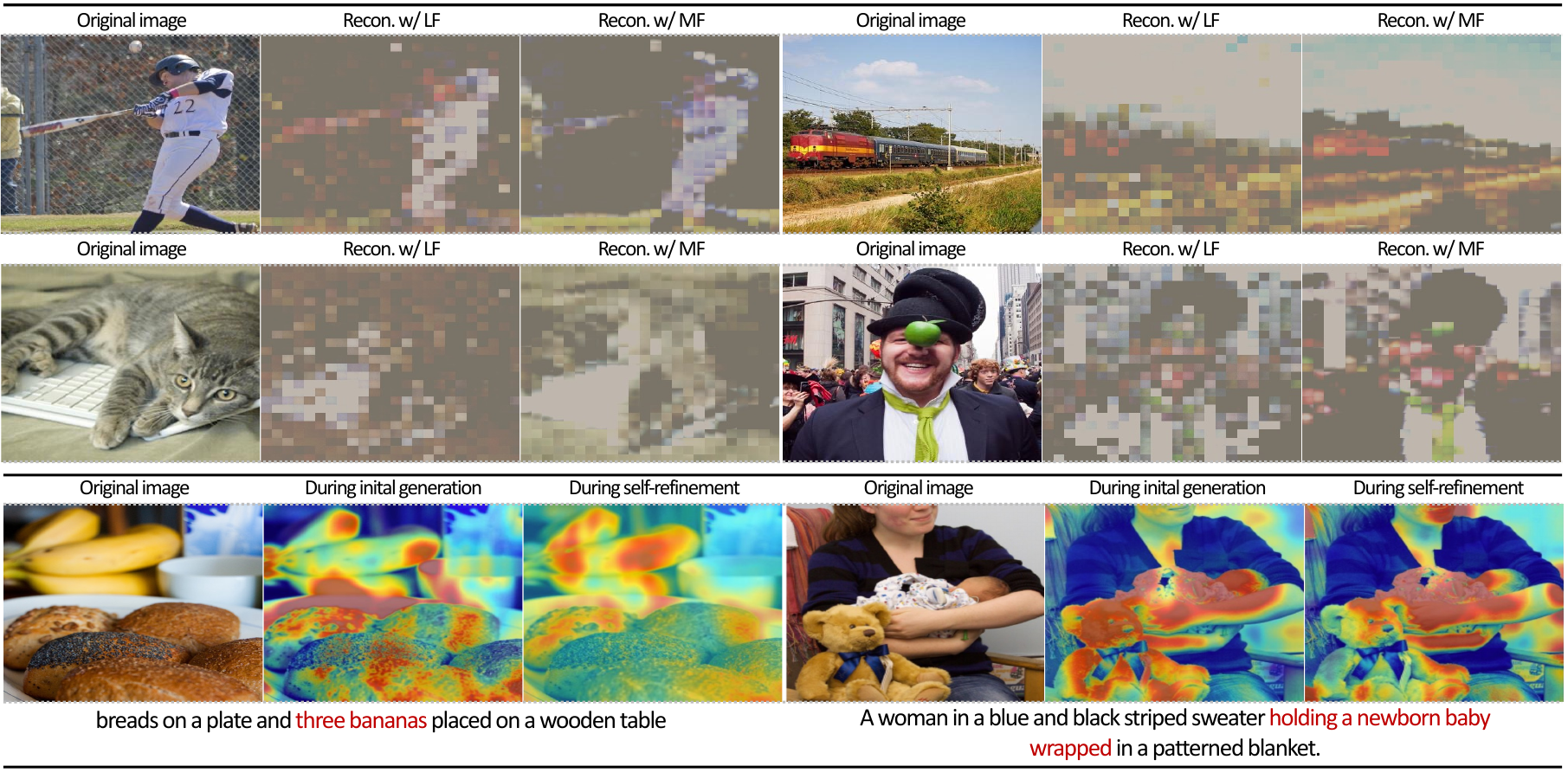}
\vspace{-2.em}
\caption{Additional results corresponding to the analyses in \secref{sec:results}. It illustrates diffuse attention patterns (top) and the limited visual detail captured by CLIP features (bottom).}
\label{fig:analysis2}
\vspace{-.5em}
\end{figure*}

\section{Supplementary Results}
\noindent\textbf{Analysis of captioning operation.}
\label{sec:added_visual}
We present additional results from the analysis experiments conducted in \secref{sec:explore}, as shown in \figref{fig:analysis2}.
In our supplementary analysis of the captioning operation, we further examine whether CLIP’s visual representations contain sufficiently fine-grained information. To this end, we adopt the decoder architecture from Masked Autoencoder (MAE)~\cite{mae}, freeze the CLIP encoder, and train only the MAE decoder on MS-COCO images. The reconstructed outputs show that images generated from the CLIP embeddings deviate noticeably from the originals and lack visual clarity. This suggests that the final-layer CLIP features are relatively coarse, reinforcing our motivation for leveraging multi-level representations to support more detailed and faithful caption generation.

\vspace{+.3em}\noindent\textbf{Comparison between our specialist and LLaVA-1.5-7B.}
\label{sec:poor_llava}
Additional comparisons between our lightweight specialist and the large multimodal generalist LLaVA-1.5-7B, discussed in \secref{sec:explore}, are provided in \figref{fig:gen_quali2}.
Despite its smaller parameter size and frozen vision encoder, our specialist delivers unexpectedly strong performance on captioning benchmarks, suggesting that compact models can still handle detailed captioning tasks effectively.

\begin{figure*}[t]
\centering
\vspace{-1em}
\begin{minipage}[t]{.77\linewidth}
    \centering
    \includegraphics[width=1.\linewidth]{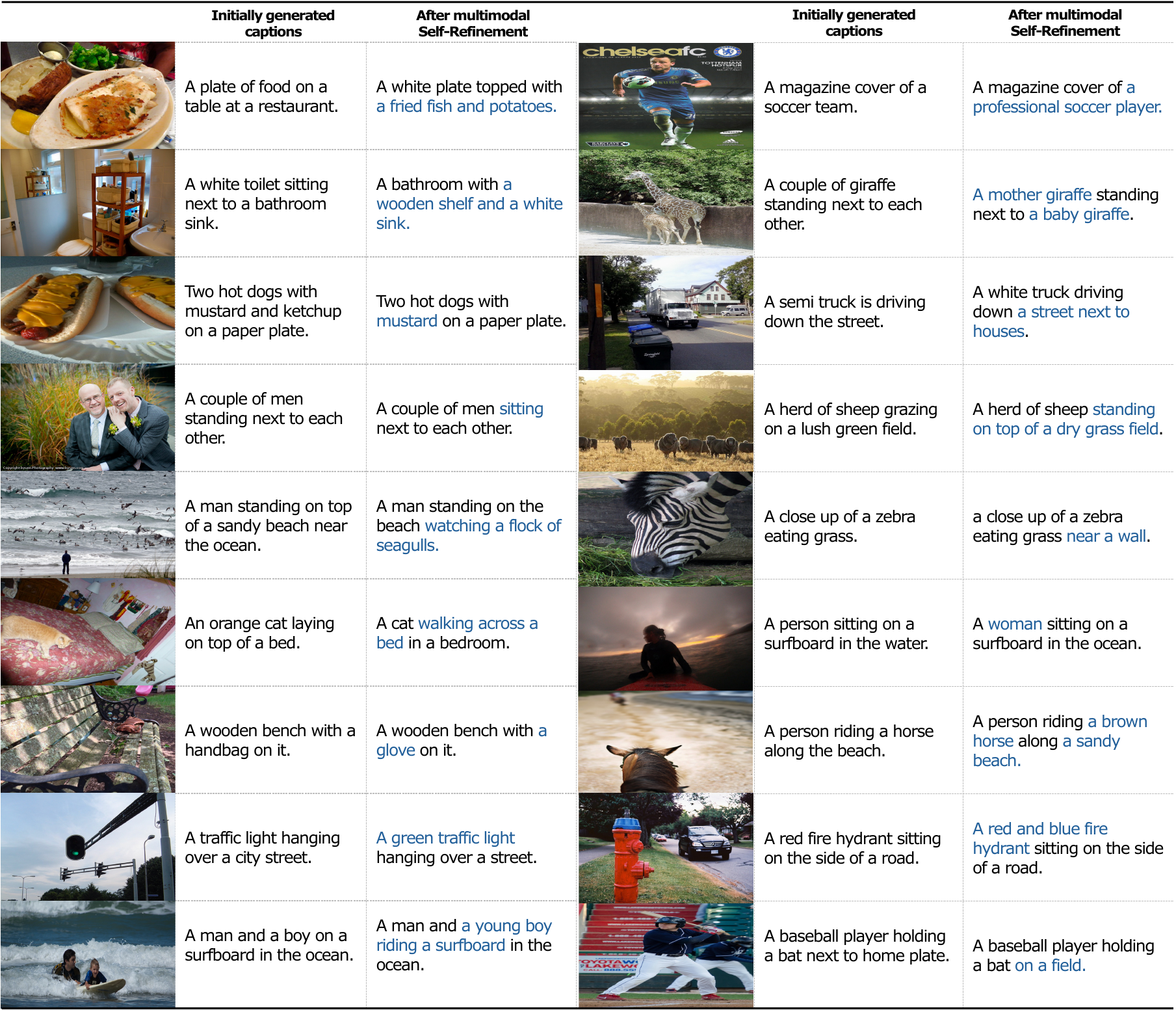}
    \vspace{0.1em}
\end{minipage}
\hfill
\begin{minipage}[t]{.77\linewidth}
    \centering
    \includegraphics[width=1.\linewidth]{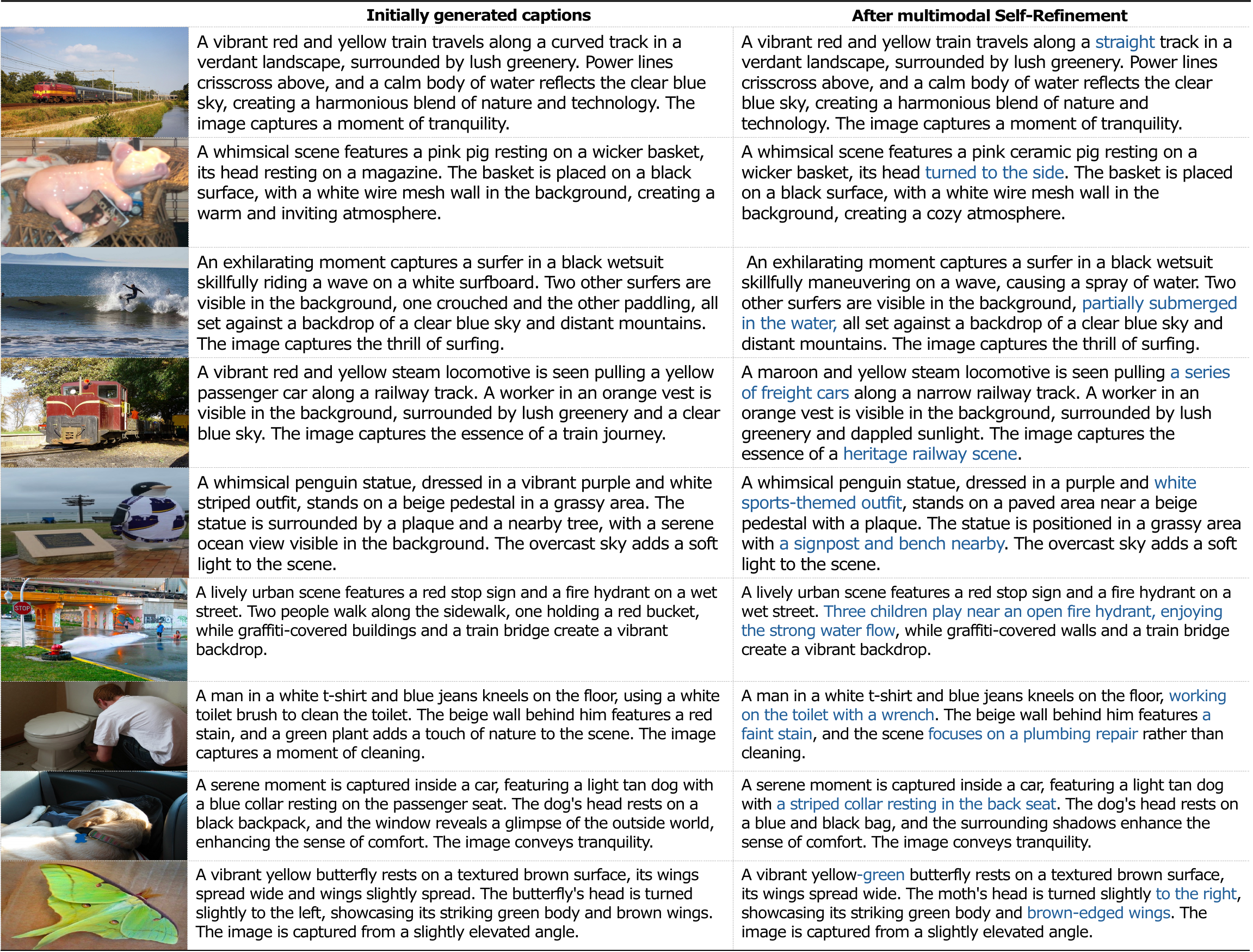}
\end{minipage}
\caption{Qualitative examples of our multimodal self-refinement. The results show improved caption quality after refinement, using MS-COCO~\cite{mscoco} (top) and ShareGPT4V~\cite{chen2024sharegpt4v} and DCI~\cite{dci} (bottom).}
\label{fig:ref_quali2}
\vspace{-1em}
\end{figure*}

\begin{figure*}[t]
\centering
\vspace{-1em}
\begin{minipage}[t]{.77\linewidth}
    \centering
    \includegraphics[width=1.\linewidth]{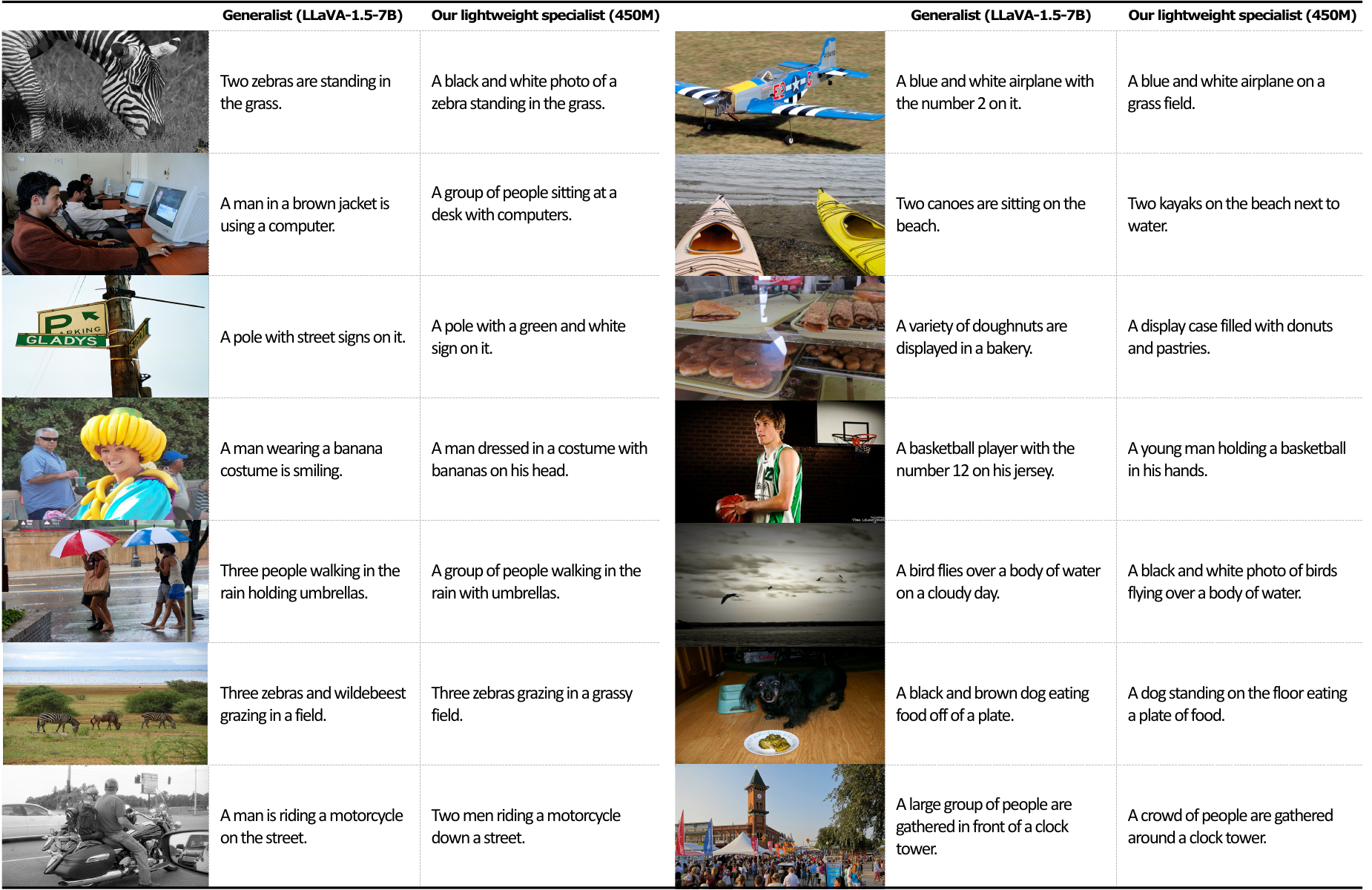}
    \vspace{0.1em}
\end{minipage}
\hfill

\begin{minipage}[t]{.77\linewidth}
    \centering
    \includegraphics[width=1.\linewidth]{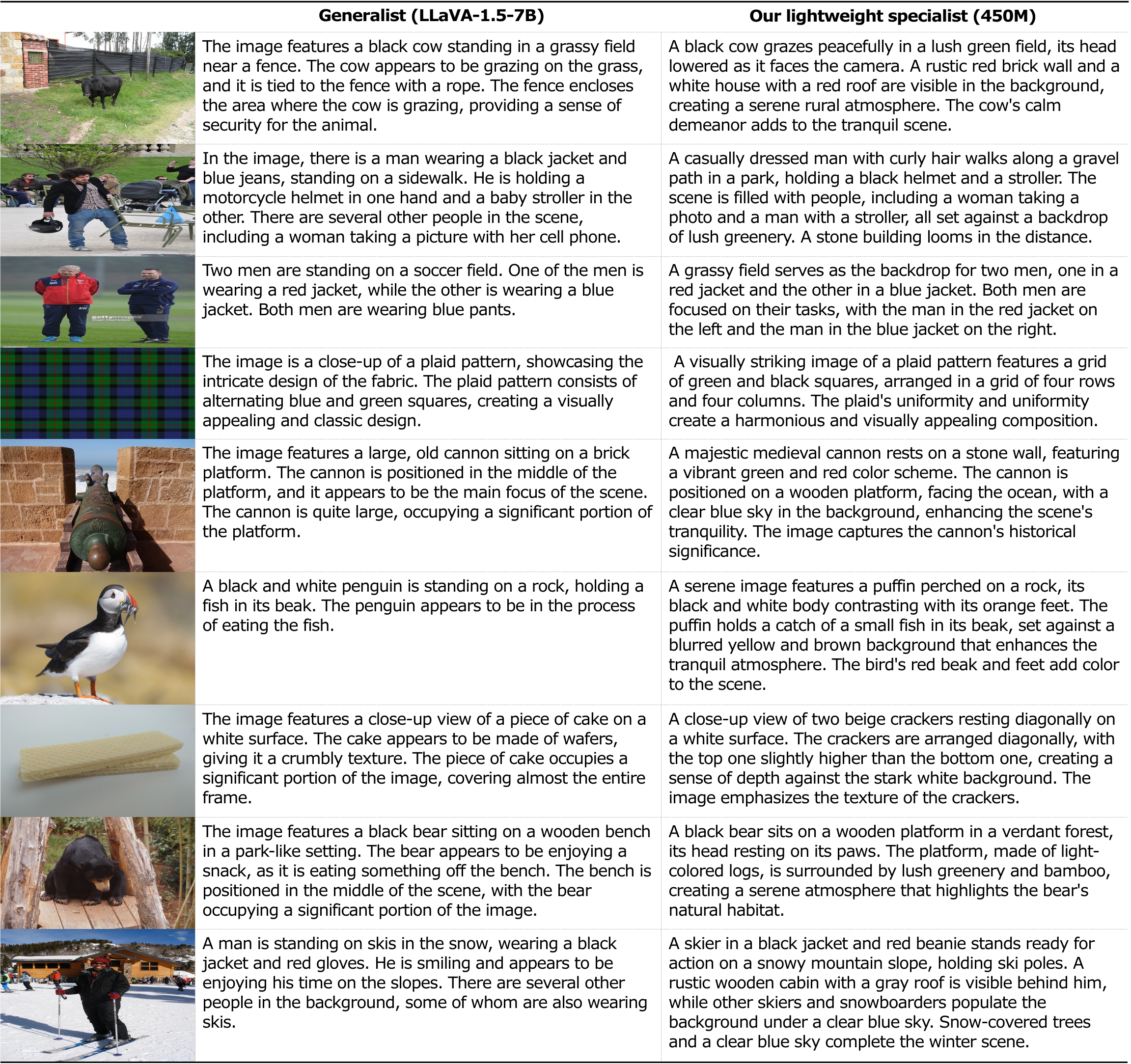}
\end{minipage}
\caption{Qualitative comparison between our lightweight specialist and the large multimodal generalist LLaVA-1.5-7B using MS-COCO~\cite{mscoco} (top) and ShareGPT4V~\cite{chen2024sharegpt4v} \& DCI~\cite{dci} (bottom). Despite its smaller size and simpler architecture, our model produces competitive descriptions.}
\label{fig:gen_quali2}
\vspace{-1em}
\end{figure*}

\vspace{+.3em}\noindent\textbf{Impact of multimodal self-refinement.}
\label{sec:added_refinement}
Additional qualitative results related to our refinement framework are shown in \figref{fig:ref_quali2}, supplementing the findings in \secref{sec:result}.
Our refinement method provides noticeable improvements in caption quality for both single-sentence and detailed captioning.

\section{Extensive Details on The Dataset}

\subsection{Pseudo-Initial Caption Generation}
\label{sec:strategy_hat_c}

We describe our strategy for generating pseudo-initial captions as follows. The instruction prompt used for this process is shown in \figref{fig:prompt2}.

\begin{itemize}[leftmargin=15pt, itemsep=1.5pt]
\item Pseudo-initial captions are created by modifying 0–3 elements of the ground-truth (GT) caption.
\item Modifications fall into four categories: Entity (e.g., chair, cat), Attribute (e.g., color, material such as wooden, count such as three cups, texture, shape, size, inspired by DSG~[1]), Relation (e.g., A in front of B), and Action (e.g., eating, blowing).
\item Apart from these modifications, the overall sentence structure and style should be preserved.
\item The pseudo-initial caption may occasionally be identical to the GT caption.
\item Few-shot examples are provided, as shown in \tabref{tab:gt_pseudo_examples}.
\end{itemize}

\begin{table*}[t]
\tablestyle{4pt}{1.2}
\caption{Few-shot examples of GT captions and corresponding pseudo-initial captions.}
\vspace{-1em}
\resizebox{0.95\linewidth}{!}{%
\begin{tabular}{x{185} x{185}}
\shlinesmall
\hdashline
\rowcolor[gray]{0.95}
GT caption & pseudo-initial caption \\
\hdashline
\multirow{3}{*}{A woman in a room with a cat}
  & A woman in a room with two cats \\
  & A woman in a room with a dog \\
  & A woman in a room with a cat \\
\hdashline
\multirow{3}{*}{A multicolored motorcycle rests outside of a sheep farm}
  & A multicolored bicycle rests inside a sheep farm \\
  & A bright red motorcycle rests outside of a sheep farm \\
  & A multicolored motorcycle races around a sheep farm \\
\shlinesmall
\end{tabular}
}
\label{tab:gt_pseudo_examples}
\end{table*}

We consistently generate three pseudo-initial captions per GT caption. Hence, the dataset can be summarized as \tabref{tab:num_peudo}, where ‘\#pairs’ indicates the number of image-caption pairs. 

\begin{table*}[h]
\tablestyle{4pt}{1.2}
\caption{Statistics of datasets used for fine-tuning. Notably, the number of images in fine-tuning stages 1 and 2 are \textit{identical}.}
\vspace{-1em}
\resizebox{0.95\linewidth}{!}{%
\begin{tabular}{x{45} x{50} x{65} x{65} x{80} x{65}}
\shlinesmall
\hdashline
\rowcolor[gray]{0.95}
Dataset & \#images & \#GT captions per image & \#pairs for fine-tuning stage 1 & \#pseudo-initial captions per GT caption & \#pairs for fine-tuning stage 2 \\
\hdashline
COCO~\cite{mscoco}       & 113K  & 5  & 565K & 3 & 1.6M \\
ShareGPT4V~\cite{chen2024sharegpt4v} & 100K  & 5  & 500K & 3 & 1.5M \\
DCI~\cite{dci}        & 7.4K  & 10 & 74K  & 3 & 0.2M \\
GLaMM~\cite{rasheed2024glamm}      & 550K  & 1  & 550K & 3 & 1.6M \\
\shlinesmall
\end{tabular}
}
\label{tab:num_peudo}
\end{table*}




\subsection{Datasets for Detailed Captioning}
\label{sec:detailed_dataset}
While MS COCO~\cite{mscoco} has long been the standard benchmark for image captioning, its single-sentence annotations often fail to capture the full richness of visual content. Recent works~\cite{ye2025painting, gabbay2021zerodim, li2023factual, dong2024benchmarking} highlight this limitation and underscore the need for datasets that support more detailed captioning.
In this study, we use three datasets that provide higher-quality and more comprehensive image descriptions: ShareGPT-4V~\cite{chen2024sharegpt4v}, DCI~\cite{dci}, and GLaMM~\cite{rasheed2024glamm}. ShareGPT-4V captions are initially generated by GPT-4o and subsequently refined by human annotators. DCI consists of fully human-written captions. GLaMM, in contrast, produces detailed descriptions by combining outputs from multiple open-source tools, including object detectors and scene parsers, and composing them using an LLM.

\subsection{ShareGPT \& DCI}
\label{sec:sharegpt_details}
The DCI dataset~\cite{dci} contains 7.4K training images and 0.4K test images, each paired with 10 human-written captions averaging 55 words.
The ShareGPT4V dataset~\cite{chen2024sharegpt4v} includes 100K images, with each image accompanied by a single long, human-verified caption of roughly 200 words. Directly using this format poses challenges for n-gram–based evaluation metrics, which benefit from multiple reference captions per image. To address this, we summarize each original caption into five shorter captions of approximately 50 words. Generating multiple long captions risks introducing hallucinations, whereas summarization preserves the original content faithfully. The prompt used for this process is shown in \figref{fig:prompt1}, and the resulting dataset will be made publicly available.
Because DCI is relatively small compared to MS COCO (118K images with five captions each), we combine DCI with the processed ShareGPT4V dataset to create a unified benchmark. This yields 102.4K training images (100K from ShareGPT4V and 7.4K from DCI), each with five or ten detailed captions, and 5K test images—making the combined dataset comparable in scale to MS COCO.

\subsection{GLaMM}
The GLaMM dataset~\cite{rasheed2024glamm} contains automatically generated captions produced using a combination of object detection models, scene-graph parsers, and LLMs. Each caption is approximately 45 words on average. In our experience, despite leveraging a wide range of visual tools, the caption quality is often inconsistent. We observed frequent factual inaccuracies, typically one or two incorrect words appearing every two to three captions.
Each image in GLaMM is paired with a single caption, which poses challenges for n-gram–based evaluation, as previously discussed. Nevertheless, we include this dataset in our experiments. We randomly sample 600K image–caption pairs from the full dataset, allocating 30K for testing and using the remaining 570K for training. Possibly due to the quality issues noted above, models trained on GLaMM generally underperform compared to those trained on ShareGPT4V and DCI.

\section{Prompt Templates}
\label{sec:prompt}

\begin{figure*}[t]\centering
\includegraphics[height=0.8\textheight, keepaspectratio]{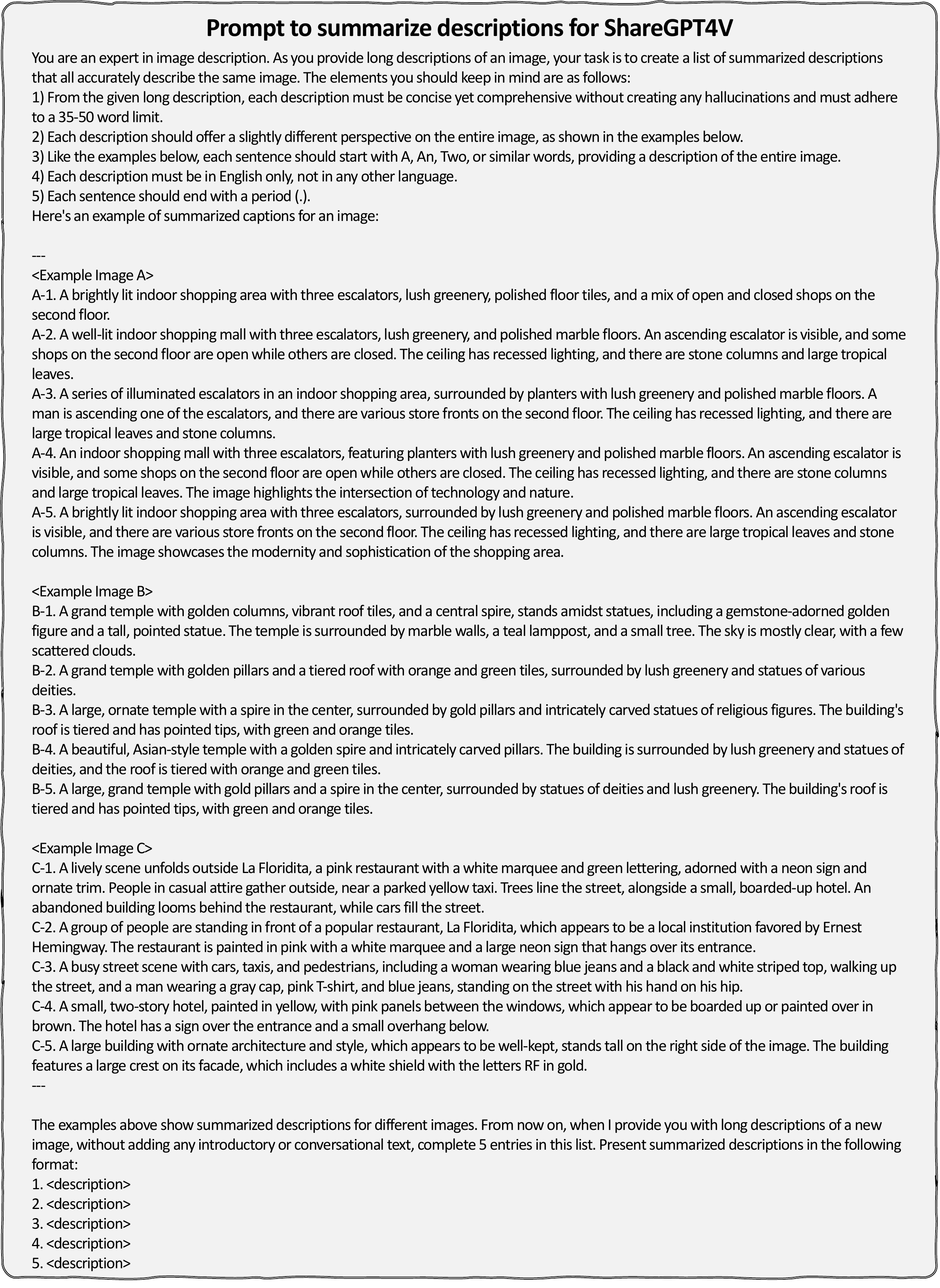}
\vspace{-1.em}
\caption{Prompt used for summarizing long-form captions into shorter, multi-reference captions.}
\label{fig:prompt1}
\vspace{-1.em}
\end{figure*}

\begin{figure*}[t]\centering
\includegraphics[width=1.\linewidth, keepaspectratio]{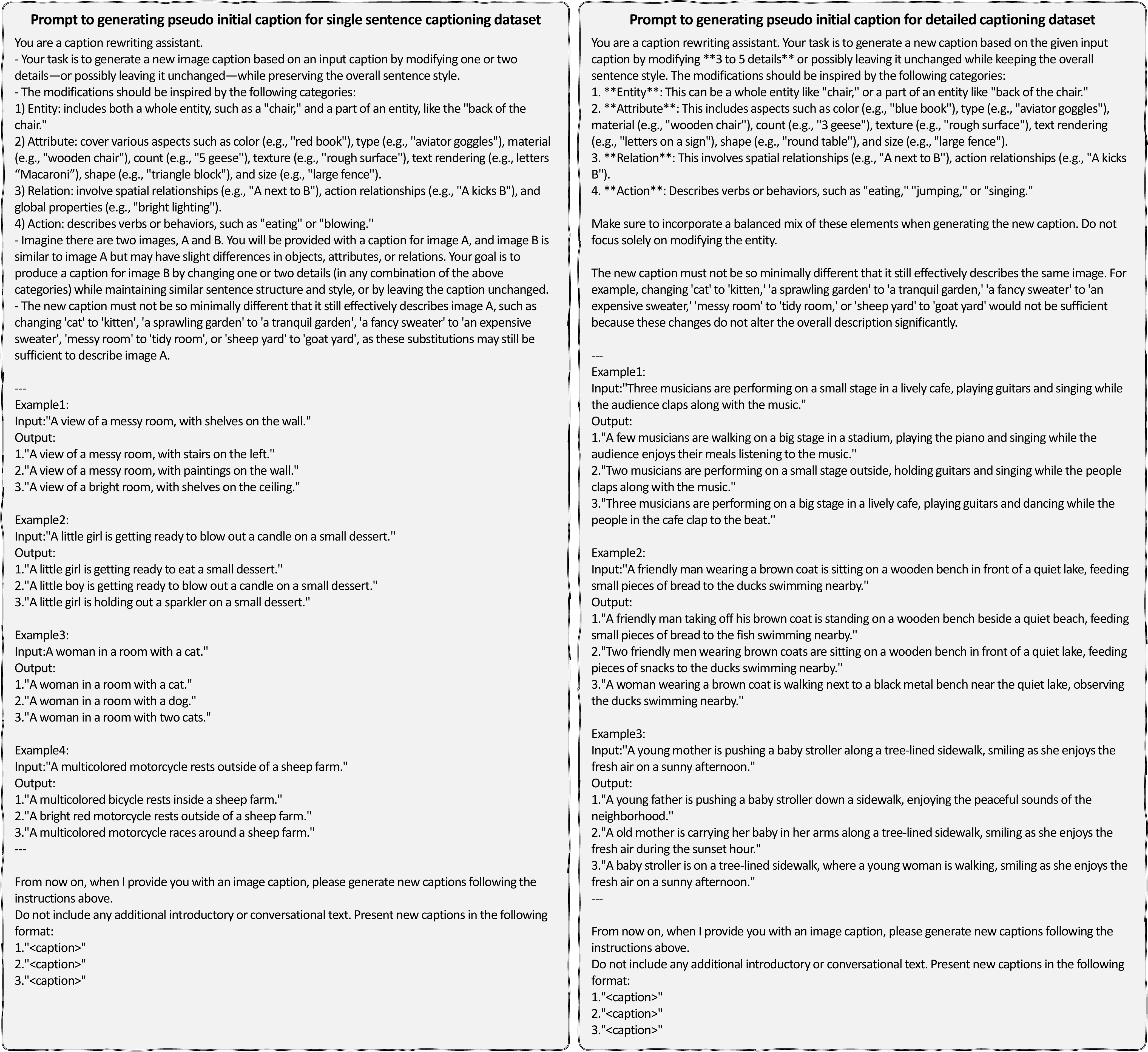}
\vspace{-1.em}
\caption{Prompt used to generate pseudo-initial captions for \MSeR.}
\label{fig:prompt2}
\end{figure*}

\begin{figure*}[t]\centering
\includegraphics[height=1\textheight, keepaspectratio]{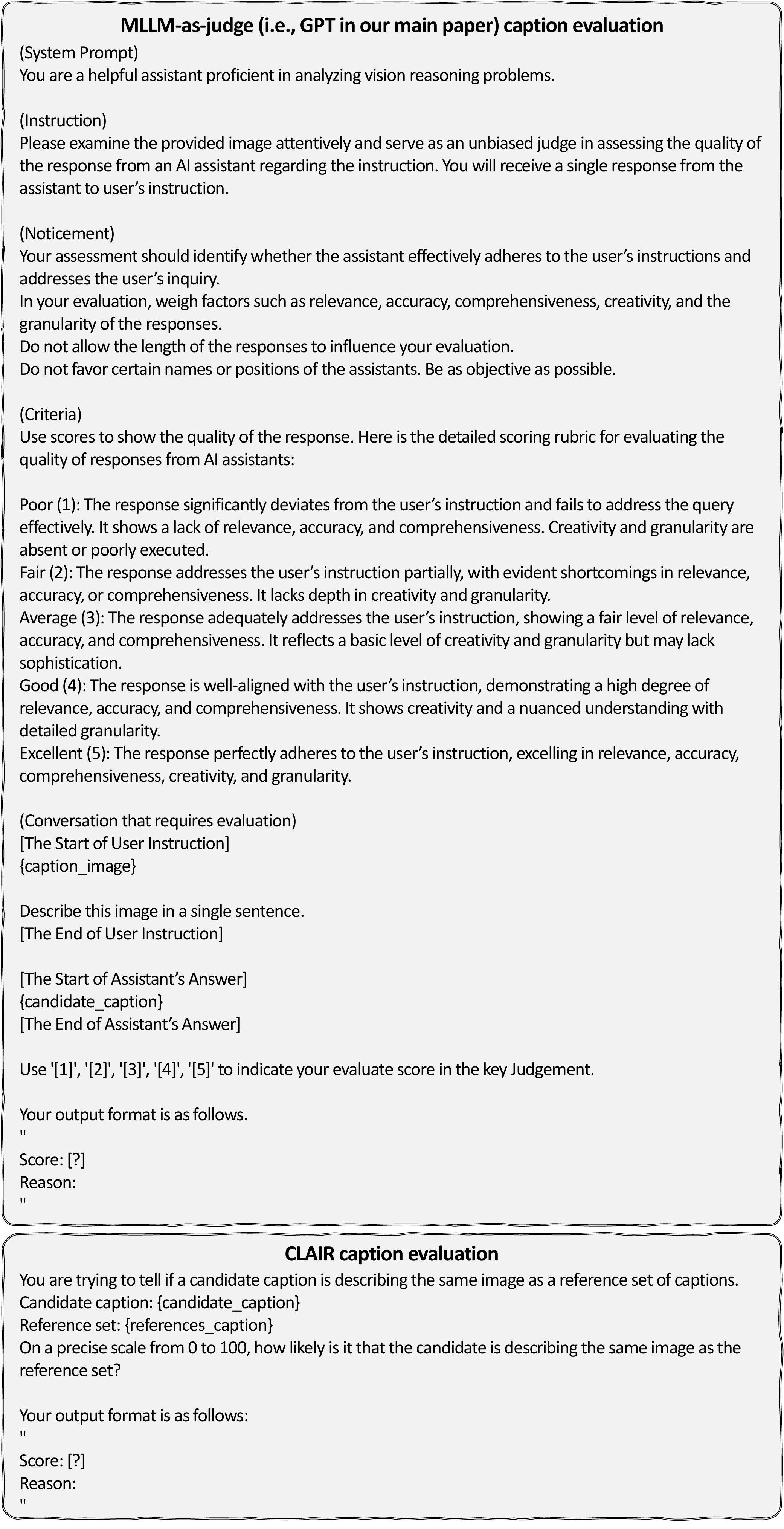}
\vspace{-1.em}
\caption{Prompt used for MLLM-as-judge evaluation, following the original template from~\cite{clair,mllmjudge}.}
\label{fig:prompt3}
\end{figure*}

\begin{figure*}[t]\centering
\includegraphics[width=1.\linewidth]{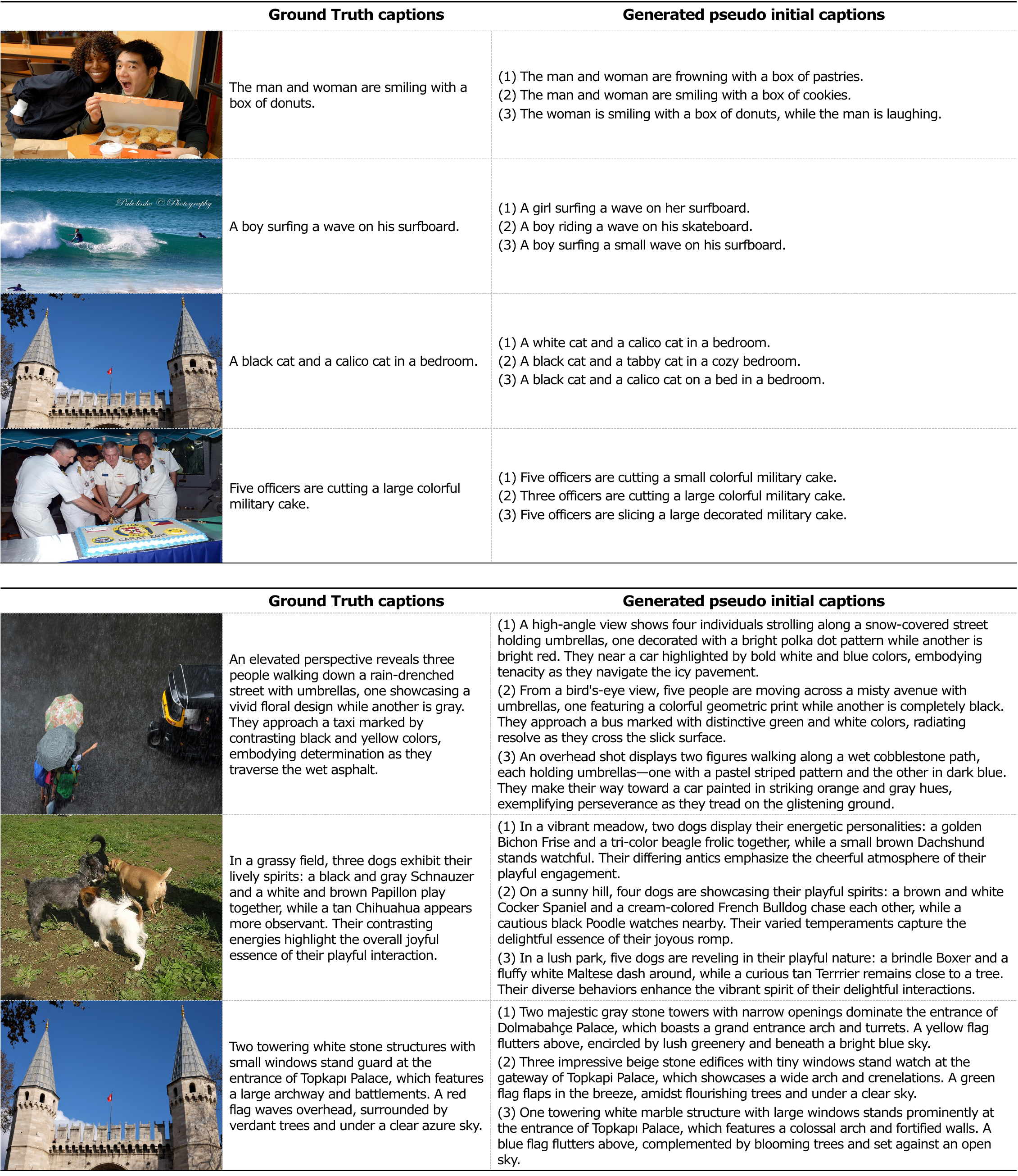}
\vspace{-2.em}
\caption{Examples of pseudo initial captions generated using the prompt in \figref{fig:prompt2}, from MS COCO (top) and ShareGPT4V \& DCI (bottom).}
\label{fig:prompt2_out}
\vspace{-2.em}
\end{figure*}

We provide the prompt templates used when interacting with OpenAI’s GPT models throughout our study. The prompt for summarizing long captions into shorter ones, introduced in \secref{sec:sharegpt_details}, is shown in \figref{fig:prompt1}. The prompt used to generate pseudo-initial captions for our refinement framework is presented in \figref{fig:prompt2}, with example outputs shown in \figref{fig:prompt2_out}. Finally, the prompt used in the MLLM-as-judge evaluation setup is provided in \figref{fig:prompt3}, where we closely follow the template proposed in the original MLLM-as-judge papers~\cite{clair, mllmjudge}.





\section{Experimental Details}
\subsection{Pretraining and Finetuning}
\label{sec:added_implementation}

Our implementation is based on the LLaVA-1.5 repository~\cite{llava15}. To reduce reliance on large language models, we replace LLaMA with the OPT series~\cite{zhang2022opt}. Training proceeds in three stages: pretraining, finetuning for caption generation (described in \secref{sec:explore}), and finetuning for multimodal self-refinement (described in \secref{sec:training}). The hyperparameters used in each stage are summarized in \tabref{tab:pretrain_finetune}. We closely follow the original LLaVA training configuration. As highlighted in \secref{sec:whyexisting}, our method differs from prior small-model approaches in that visual features are injected directly into the self-attention inputs of the language model.

\begin{table*}[t]
\tablestyle{4pt}{1.2}
\caption{Hyperparameters used for model training. The settings to the left and right of the \textcolor{darkred}{\textbf{/}} correspond to those used in \secref{sec:glance} and \secref{sec:explore}, respectively.}
\vspace{-1em}
\resizebox{0.9\linewidth}{!}{%
\begin{tabular}{x{80} ;{.5pt/1.pt} x{150} ;{.5pt/1.pt} x{150}}
\shlinesmall
\rowcolor[gray]{0.95}
 & Pretraining & Fine-tuning \\
\hdashline
Dataset & LCS-558K~\cite{llava15} & MS COCO~\cite{mscoco} or SharedGPT~\cite{chen2024sharegpt4v} + DCI~\cite{dci} \\
\hdashline
Adapter & 4-layer MLP with GELU & 4-layer MLP with GELU \textcolor{darkred}{\textbf{/}} Deeplens \\
\hdashline
Trainable & Adapter layers only & Adapters + Language Model \\
\hdashline
Training Epochs & 1 & 10 \textcolor{darkred}{\textbf{/}} 2 \\
\hdashline
Learning Rate & $1 \times 10^{-4}$ & $2 \times 10^{-5}$ \\
\hdashline
Weight Decay & 0 & 0 \\
\hdashline
Warm-up Ratio & 0.03 & 0.03 \\
\hdashline
Learning Rate Scheduler & Cosine decay & Cosine decay \\
\shlinesmall
\end{tabular}
}
\label{tab:pretrain_finetune}
\end{table*}

\subsection{Experiment setup of MLLMs}
\label{sec:generalist-prompt}
\tabref{tab:gen_perform} compares specialist models with several generalist MLLMs, including InstructBLIP~\cite{dai2023instructblip}, Unified-IO-XL~\cite{lu2023unifiedio}, Shikra~\cite{chen2023shikra}, Qwen-VL~\cite{bai2023qwenvl}, and LLaVA-1.5~\cite{llava15}. Although these generalist models were trained on MS COCO images, they were not trained on datasets such as ShareGPT4V, DCI, or GLaMM. Instead, they were instruction-tuned on large-scale multimodal datasets; for example, Qwen-VL and InstructBLIP were trained on approximately 1.5B and 130M instruction samples, respectively. For this reason, we categorize them as generalist models.
We evaluate the generalist MLLMs using publicly available checkpoints from their official repositories, without additional fine-tuning. For the single-sentence captioning task, we use the prompt:
\textit{Provide a one-sentence caption for the provided image.''} For the detailed captioning task, we use the prompt: \textit{Describe the photo within 55 words.''}
We also include an additional evaluation of LLaVA-1.5 using an alternative instruction, as shown in \tabref{tab:llava_prompts}.
These results highlight two key observations: (1) evaluation metrics such as BERTScore~\cite{bertscore} tend to penalize long-form outputs, and (2) longer generations increase hallucination frequency, consistent with prior findings~\cite{jiang2024hallucination}.

\begin{table}[t]
\tablestyle{4pt}{1.2}
\caption{Performance of LLaVA-1.5 under different prompt instructions.}
\vspace{-1em}
\resizebox{1\linewidth}{!}{%
\begin{tabular}{x{110} x{35} x{45} x{35}}
\shlinesmall
\hdashline
\rowcolor[gray]{0.95}
LLaVA-1.5 & CIDEr~\cite{vedantam2015cider} & BERTScore~\cite{bertscore} & CAPT~\cite{dong2024benchmarking} \\
\hdashline
``Describe the photo within 55 words'' & 36.1 & 36.6 & 48.6 \\
``Describe the photo in detail''        & 12.8 & 17.6 & 40.6 \\
\shlinesmall
\end{tabular}
}
\label{tab:llava_prompts}
\end{table}

\subsection{Attention map visualization}
\label{sec:added_implementation_attention}
As part of our analysis in \secref{sec:results}, we evaluate whether the model attends to the appropriate image regions when generating specific words in a caption. To this end, we adapt the visualization code provided by API~\cite{yu2024attention}, originally designed to highlight attention maps between images and questions in VQA tasks.
We modify the attention hooking module\footnote{\url{https://github.com/yu-rp/apiprompting/blob/master/API/API_LLaVA/hook.py}} to visualize attention between image regions and selected words within captions. This enables us to examine which areas the language model focuses on when producing particular tokens.

\subsection{Image reconstruction}
\label{sec:added_implementation_recon}
To investigate whether CLIP’s visual representations are coarse or ambiguous, we conducted an image reconstruction experiment using a Masked AutoEncoder (MAE) framework (details in \secref{sec:glance}). In this setup, a Vision Transformer (ViT) encoder produces visual features, which the decoder subsequently uses to reconstruct the image.
We adopt the same visual encoder as used in LLaVA-1.5, \texttt{CLIP ViT-L/14-336}, and keep its parameters frozen. The decoder receives the visual embeddings without masking and predicts the corresponding RGB image. We utilize two types of visual inputs: (i) Last-layer Features (LF), and (ii) Multi-Level Features (ML), where ML consists of outputs from layers 13, 18, and 23 of the encoder.
The decoder was trained on 100K images from the MS COCO dataset. Further architectural details are available in the MAE repository\footnote{\url{https://github.com/facebookresearch/mae/blob/main/models_mae.py}}.



\subsection{Long Range Video Question Answering}
\label{sec:added_implementation_videoqa}

We evaluate our model on the Long-Range Video Question Answering task~\cite{zhang2023simple}. This task requires the system to answer multiple-choice questions based on user queries about videos that are 10 minutes or longer in duration. We emphasize that currently, no video MLLMs (Multimodal Large Language Models) are capable of handling this task directly. Most existing models~\cite{li2024llava, bai2025qwen2, zhu2025internvl3} impose limits on the number of input video frames they can process, making it difficult to cover the full temporal span of such long videos.
To address this limitation, recent approaches~\cite{ye2025re, wang2024videoagent} propose first extracting per-frame captions. Subsequently, both the generated captions and the question are injected into an LLM capable of processing over 100K tokens in a single prompt.
For evaluation, we follow the setup provided in the official implementation of LLoVi\footnote{\url{https://github.com/CeeZh/LLoVi}}. replacing only the captioning models.
(i) When using the generalist model, we generate a caption for each frame using the prompt: "Describe this frame within 50 words.". (ii) 
For our specialist model, we utilize the same prompt and apply the version trained on the ShareGPT4V~\cite{chen2024sharegpt4v} and DCI~\cite{dci} datasets.

\end{document}